\documentclass[twoside,11pt]{article}

\usepackage{blindtext}

\usepackage[nohyperref,preprint]{jmlr2e}

\usepackage{microtype}
\usepackage{graphicx}
\usepackage{caption}
\usepackage{subcaption}
\usepackage{booktabs} 

\usepackage{xcolor}
\definecolor{darkblue}{rgb}{0,0,0.5}
\definecolor{firebrick}{rgb}{0.75,0.125,0.125}
\definecolor{darkgreen}{rgb}{0,0.5,0}
\definecolor{light-gray}{gray}{0.5}

\definecolor{grid_search}{HTML}{1f77b4}
\definecolor{ax}{HTML}{ff7f0e}
\definecolor{grid_search}{HTML}{1f77b4}
\definecolor{ax}{HTML}{ff7f0e}
\definecolor{hebo}{HTML}{2ca02c}
\definecolor{blend_search}{HTML}{9467bd}
\definecolor{bayesopt}{HTML}{8c564b}
\definecolor{skopt}{HTML}{e377c2}
\definecolor{nevergrad}{HTML}{7f7f7f}
\definecolor{random}{HTML}{d62728}
\definecolor{hyperopt}{HTML}{bcbd22}
\definecolor{optuna}{HTML}{17becf}
\definecolor{bohb}{HTML}{aec7e8}
\definecolor{cfo}{HTML}{98df8a}
\definecolor{zoopt}{HTML}{ff9897}

\usepackage[colorlinks=true,linkcolor=firebrick,citecolor=darkgreen,urlcolor=darkblue]{hyperref}
\usepackage{url}
\usepackage{units}

\usepackage{notations}
\usepackage{caption}
\usepackage{graphicx}
\graphicspath{ {./figures/} }
\usepackage[textsize=tiny]{todonotes}

\usepackage{lastpage}
\jmlrheading{24}{2023}{1-\pageref{LastPage}}{1/21; Revised 5/22}{9/22}{21-0000}{Bal\'azs K\'egl}

\ShortHeadings{A systematic study comparing hyperparameter optimization engines on tabular data}{A systematic study comparing hyperparameter optimization engines on tabular data}
\firstpageno{1}

\begin{document}

\title{A systematic study comparing hyperparameter optimization engines on tabular data}

\author{
    \name Bal\'azs K\'egl
    \addr Noah's Ark Lab, Huawei Paris
    \email balazs.kegl@gmail.com
}

\editor{My editor}

\maketitle

\begin{abstract}
We run an independent comparison of all hyperparameter optimization (hyperopt) engines available in the Ray Tune library. We introduce two ways to normalize and aggregate statistics across data sets and models, one rank-based, and another one sandwiching the score between the random search score and the full grid search score. This affords us i) to rank the hyperopt engines, ii) to make generalized and statistically significant statements on how much they improve over random search, and iii) to make recommendations on which engine should be used to hyperopt a given learning algorithm. We find that most engines beat random search, but that only three of them (HEBO, AX, and BlendSearch) clearly stand out. We also found that some engines seem to specialize  in hyperopting certain learning algorithms, which makes it tricky to use hyperopt in comparison studies, since the choice of the hyperopt technique may favor some of the models in the comparison.
\end{abstract}

\begin{keywords}
  hyperparameter optimization, hyperopt, Ray Tune, tabular data
\end{keywords}

%

\section{Introduction}
\label{secIntroduction}

Sequential gradient-free (Bayesian) optimization has been the go-to tool for hyperparameter optimization (hereafter \emph{hyperopt}) since it was introduced to machine learning by \citet{bergstra2011} and \citet{snoek2012}. The basic algorithmic skeleton is relatively simple: we learn a probabilistic  surrogate on the finished trials then we optimize an acquisition function to propose a new arm (hyperparameter value combination). That said, there are numerous  bricks and heuristics that afford a large space of algorithmic variants, and choosing the right hyperopt technique and library is challenging for a practicing data scientist.

Hyperparameter optimization is a crucial step in designing well-performing models \citep{zhang2021}. It is also very expensive since every search step requires the training and scoring a learning algorithm. Comparing hyperopt engines is thus doubly expensive, since we need to run many hyperopt experiments to establish statistical significance between the performance of the engines. Not surprisingly, only a handful of papers attempt such comparison \citep{falkner2018,li2018,klein2019,cowen2020, eggensperger2021}, and many times these studies are not independent: they introduce a new technique which is also part of the comparison. It is also common to use tiny UCI data whose relevance to real-world data science is questionable.

Integration libraries such as \href{https://scikit-learn.org/stable/modules/generated/sklearn.ensemble.RandomForestClassifier.html}{\textsc{ScikitLearn}} \citep{pedregosa2011},  \href{https://docs.ray.io/en/latest/tune/index.html}{Ray Tune}~\citep{liaw2018}, and \href{https://www.openml.org/}{\textsc{OpenML}} \citep{vanschoren2013,feurer2019} provide unified APIs to machine learning models, data, and hyperopt engines, respectively. Ray Tune is especially  useful as a model selection and hyperparameter optimization library that affords a unified interface to many popular hyperparameter optimization algorithms we call \emph{engines} in this paper. Section~\ref{secEngines} is dedicated to a brief enumeration of all the engines participating in this comparison.

Thanks to these awesome tools, hyperopt comparison can be done relatively painlessly. In this paper, we put ourselves into the shoes of a practicing data scientist who is not an expert of hyperopt, and would just like to choose the best engine with default parameters from a library that provides a unified API to many engines. We aim at answering basic questions: by running a hyperopt engine versus random search, i) how much do I gain on the score and ii) how much time do I save? This is an experimental integration study, which means that we do not introduce any new hyperopt technique. That said, we believe that our results point to the right directions for hyperopt researchers to improve the techniques.

Our basic methodology is to first run grid search on a coarse predefined search grid, essentially pre-computing all the scores that the various engines with various seeds ask for in their attempt to find the optimum. This means that we decouple the expensive train-test-score step from the sequential optimization, affording us space to achieve statistical significance. This choice also means that we restrict the study to a grid search (with limited budget), which favors some of the engines. We have good arguments (Section~\ref{secWhyGrid}) to support this decision, not only because it makes the comparison possible, but also because it is a good practice.

Another constraint that comes with the requirement of running many hyperopt experiments is that we need to limit the size of the training data sets. As with the grid constraint, we argue that this does not make the study ``academic'': most of the data in the real world are small. In our experiments (Section~\ref{secResults}) we use ten-fold cross validation on 5000 data points, uniformly across data sets and models. This is small enough to run a meaningful set of experiments and big enough to obtain meaningful results. On the other hand, to eliminate test variance, we select data sets that afford us huge test sets, thus precise measurements of the expected scores.

One of the problems we need to solve for establishing statistically significant differences is aggregation: we need to be able to average results across metrics, data sets, and models. We introduce two metrics to solve this problem. Rank-based metrics (Section~\ref{secRankMetrics}) answer the question: what is the probability that an engine performs better than random search? We design a statistics based on the discounted cumulative gain metrics to answer this question. Score-based metrics (Section~\ref{secScoreMetrics}) answer the question: how much do we improve the score of random search by using a hyperopt engine? The issue here is the different scale of scores across metrics, data sets, and models, which we solve by sandwiching the scores between those of random search and grid search. Averaging rank-based metrics is more proper, but score-based metrics measure the quantity that we want to optimize in practice. 

Our study is also limited to a purely sequential protocol. While we tend to agree that distributing hyperopt is the best way to accelerate it, adding distributedness to the comparison study raises several methodological questions which are hard to solve. We also do not test advanced features such as pruning and dynamically constructing hyperparameter spaces. These are useful when dealing with complex hyperopt problems, but these are hard to compare systematically, and they fall out of the scope of this paper. We would add tough that even when we use these sophisticated heuristics, search in a fixed space lies at the heart of hyperopt, so knowing where to turn when such a step is needed leads to an overall gain.

Here is the summary of our findings.
\begin{itemize}
    \item Most engines are significantly better than random search, with the best ones accelerating the search two to three times.
    \item Out of the eleven engines tested, three stand out: Huawei's \href{https://github.com/huawei-noah/HEBO/tree/master/HEBO}{\textsc{{HEBO}}} \citep{cowen2020} that won the \href{https://bbochallenge.com/}{2020 NeurIPS BBO Challenge}; Meta's \href{https://ax.dev}{\textsc{{AX}}} \citep{bakshy2018}; and Microsoft's \href{https://github.com/microsoft/FLAML/tree/main/flaml/tune#blendsearch-economical-hyperparameter-optimization-with-blended-search-strategy}{\textsc{{BlendSearch}}} \citep{wang2021}.
    \item Some engines seem to specialize in hyperopting certain learning algorithms. This makes it tricky to use hyperopt in comparison studies, since the choice of the hyperopt technique may favor some of the models in the comparison. 
\end{itemize}










\section{The experimental methodology}
\label{secSetup}

Most machine learning prediction models $f(\bx; \btheta)$ come with a few hyperparameters $\btheta = (\theta_1, \ldots, \theta_D)$, typically $D \in \{2, \dots, 10\}$. Data scientists tune these hyperparameters to a given data set $\cD = \big\{(\bx_i, y_i)\big\}_{i=1}^n$. Each \emph{trial} $\ell$ will test a vector of hyperparameter values $\btheta^\ell$ that we will call \emph{arms} (following the multi-armed bandit terminology). Each arm $\btheta^\ell$ will be 
pulled $K=10$ times on $K$ pairs of training/validation sets $\big\{(\cD_\text{trn}^k, \cD_\text{val}^k)\big\}_{k=1}^K$ drew randomly from the data set $\cD$ ($K$-fold randomized 
cross-validation), resulting in models $f^\ell_k = \cA(\cD_\text{trn}^k, \btheta^\ell)$ where $\cA$ is a training algorithm. The trial is evaluated by an empirical validation score $\widehat{r}^\ell_k = R(f^\ell_k, \cD_\text{val}^k)$, available to the engines, and a test score $r^\ell_k = R(f^\ell_k, \cD_\text{test})$ evaluated on a held-out test set $\cD_\text{test}$, not available to the engines. 

We assume that each hyperparameter $\theta_j$ is discretized into a finite number $N_j$ of values $\theta_j \in \{v^1_j, \ldots, v^{N_j}_j\}$, for all $j=1,\ldots,D$. In this way, arms $\btheta^\ell$ are represented by the integer index vector (grid coordinate vector) $\bi^\ell = \big(i^\ell_1, \ldots, i^\ell_D\big) \in \cG$, where $\theta^\ell_j = v^{i^\ell_j}_j$ for all $j = 1,\ldots,D$, and $\cG$ is the integer grid $\cG =\prod_{j=1}^D \{1,\ldots,N_j\}$ with grid size $N =\prod_{j=1}^D N_j$.

\subsection{Why use a finite grid?}
\label{secWhyGrid}

The operational reason for using a finite grid in this study is that it lets us pre-compute the validation scores $\widehat{\cT} = \Big\{\widehat{r}^\bi_k = R\big(\cA(\cD_\text{trn}^k, \bi), \cD_\text{val}^k\big)\Big\}_{k \in \{1,\ldots,K\}}^{\bi \in \cG}$ for the full grid, letting us rapidly reading out the result when an engine pulls an arm. At the same time, we argue that pre-defining the search grid is also a good practice of the experienced data scientist, for the following reasons.
\begin{enumerate}
    \item The grid is always pre-defined by the numerical resolution, and an even smaller-resolution grid needs to be used when the acquisition function is optimized on the surrogate model (there are attempts \citep{bardenet2010}  to improve the inherent grid search of that step). So the decision is not on whether we should use grid or continuous search, but on the \emph{resolution} of the grid.
    \item Since the task is \emph{noisy} optimization ($\widehat{r}^\ell_k \not= r^\ell_k$), there is an \emph{optimal} finite grid resolution: a too coarse grid may lead to missing the optimum, but a too fine grid combined with a perfect optimizer may lead to overfitting (this is the same reason why SGD, a suboptimal optimization technique is the state-of-the-art for training neural nets). In fact, in one of our experiments it happened that, even with a coarse grid, the full grid search lead to a worse test score than a random search on a small subset of the grid. 
    \item Data scientists usually know enough to design the grid. They can use priors about length scale (smoothness of $r$ vs. $\btheta$) to adapt the grid to the hyperparameter and data size, and inform the engine about the \emph{resolution} of the search and the possibly nonlinear scale of the hyperparameters. Arguably, on a training set of a thousand points, there is no reason to test a random forest with both 100 and 101 trees.
    \item Discretization may make algorithms simpler and more robust. Gaussian processes (GP) and bandits are easier to design when the input space is discrete. In the case of a GP surrogate, \emph{its} hyperopt is more robust if the length scale is given by the grid resolution and only the noise parameter needs to be tuned.
\end{enumerate}
The only situation when discretization is restrictive is when a hyperparameter has a large range and the objective function is rough: it has a deep and thin well around the optimum. In our experience, such hyperparameters are rare. Dealing with this rare case will require several refining  meta-iterations mimicking a line search.

As a summary, we acknowledge that the coarse grids used in our experiments may be suboptimal and may also disfavor some of the engines, nevertheless, our setup is robust and informative to the real-life hyperopt practitioner.

\subsection{The sequential hyperopt loop}

The experimental design algorithm starts with an empty history $\cH^0 = \{\}$ and iterates the following three steps for $\ell = 1,\ldots,L$:
\begin{enumerate}
    \item given the history $\cH^{\ell -1}$, design an arm $\btheta^\ell = \big(v^{i^\ell_1}_1, \ldots, v^{i^\ell_D}_D\big)$ represented by $\bi^\ell = \big(i^\ell_1, \ldots, i^\ell_D\big)$;
    \item call the training and scoring algorithms to obtain $\widehat{r}^\ell$; and
    \item add the pair $h^\ell = (\bi^\ell, \widehat{r}^\ell)$ to the history $\cH^\ell = \cH^{\ell-1} \cup \{h^\ell\}$.
\end{enumerate}

The goal is to find the optimal arm $\btheta^* = \big(v^{i_1^*}_1, \ldots, v^{i_D^*}_D\big)$ represented by indices $\bi^* = \big(i^*_1, \ldots, i^*_D\big)$, and the corresponding optimal predictor $f^* = \cA(\cD, \btheta^*)$ with optimal test risk $r^* = R\big(f^*, \cD_\text{test}\big)$. In our experiments, we use the folds $k$ in two different ways. In our rank-based metrics (Section~\ref{secRankMetrics}), we use each fold as a separate single-validation experiment, iterating the hyperopt loop $K$ times and averaging the statistics over the $K$ runs. In the $k$\/th experiment, the score is thus $\widehat{r}^\ell = R\big(\cA(\cD_\text{trn}^k, \btheta^\ell), \cD_\text{val}^k\big)$. In our score-based metrics (Section~\ref{secScoreMetrics}), each trial consists in training the models and evaluating the risk on all folds, then averaging the score \emph{inside} the hyperopt loop: $\widehat{r}^\ell = \frac{1}{K} \sum_{k=1}^K R\big(\cA(\cD_\text{trn}^k, \btheta^\ell), \cD_\text{val}^k\big)$. This is a classical way to use cross-validation inside hyperopt. A third possibility, when the choice of the fold is also delegated to the engine in Step~1 (potentially pulling the same arm  multiple times for a more precise estimate of the validation risk) is the subject of a future study.

We run all our experiments with three trial budgets:\\$L_m = m \sqrt{N}$ with $m = 1~(\text{low}), 2~(\text{medium}), 3~(\text{high})$ and grid size $N = \prod_{j=1}^D N_j$.

Hyperopt engines usually train (or update) a probabilistic surrogate model on $\cH^{\ell-1}$ in each iteration $\ell$, and design $\bi^\ell$ by optimizing an acquisition function, balancing between exploration and exploitation. This simple skeleton has quite a few nuts and bolts that need to be designed and tuned (what surrogate, how to jump-start the optimization, what acquisition function, how to robustly hyperopt the surrogate model itself, just to mention a few), so the performance of the different engines vary, even if they use the same basic loop. In addition, some engines do not use surrogate models: some successful techniques are based on evolutionary search, space-filling sampling, or local search with restarts. Arguably, in the low-budget regime, the initialization of the search is more important than the surrogate optimization, this latter becoming more useful in the medium and high-budget regimes.

\subsection{How we compare engines}
\label{secCompare}

We used two tests to compare engines: rank-based and score-based. Rank-based metrics abstract away the score so they easier to aggregate between different metrics, data sets, and models; score-based metrics are closer to what the data scientist is interested in, measuring how much one can improve the score on a fixed budget or reach a certain score with a minimum budget. To aggregate experiments, care needs to be taken to normalize the improvements across metrics, data sets, and models.

\subsubsection{Rank-based metrics}
\label{secRankMetrics}

Rank-based metrics answer the question: what is the probability that an engine performs better than random search? The basic gist is first to read out, from the pre-computed full test score table $\cT = \Big\{r^\bi = R\big(\cA(\cD_\text{trn}, \bi), \cD_\text{test}\big)\Big\}^{\bi \in \cG}$, the ranks $\brho = (\rho^1, \ldots, \rho^L)$ of the test score sequence $r^1, \ldots, r^L$, generated by a given engine ($r^\ell$ is the $\rho^\ell$th best score in $\cT$). Once generated, we compare $\brho$ to a random draw $\widetilde\brho$ of $L$ ranks from the integer set $\cN = \{1, \ldots, N\}$, where $N = |\cG| = |\cT|$ is the grid size. $\widetilde\brho$ represents the rankings produced by random search with the same budget $L$. To compare $\brho$ to $\widetilde\brho$, we use a statistics $s$ designed according to what we expect from a good hyperopt engine: find good arms as fast as possible. Formally, let $s: \cN_L \rightarrow \RR$ be a function that maps $\cN_L$, a set of $L$ integers from $[1, N]$, to the real line. For simplicity, without the loss of generality, we assume that $s$ assigns higher values to better rankings. We then define
\[
 p(\text{better than random}) = p\big(s(\brho) > s(\widetilde\brho)\big),
\]
where $\widetilde\brho \in \cN$ is a random set of integers drawn without replacement form $[1, N]$. For some statistics $s$, $p$ can be computed analytically, but in our experiments we simply use $J = 10^5$ random draws $\{\widetilde\brho_j\}_{j=1}^J$ and estimate $p$ by counting the number of times that $s(\brho)$ beats $s(\widetilde\brho_j)$
\begin{equation}\label{eqnBetterThanRandom}
 p(\text{better than random}) \approx \frac{1}{J} \sum_{j=1}^J \IND{s(\brho) > s(\widetilde\brho_j)}.
\end{equation}
We experimented with various statistics, for example: the time to reach a top 10\% arm, or the bottom (best) rank in $\brho$. We report results using the \emph{discounted cumulative gain} (DCG), a popular metrics used for scoring ranking algorithms. We use DCG$_{10\%}$ which is a weighted count of top 10\% arms present in the $L$ arms generated by the engine. DCG$_{10\%}$ is a ``shaded'' statistics that favors low-rank (good) arms in $\brho$ appearing as early as possible. Formally, $\text{DCG}_{10\%}$ is defined by
\begin{equation}\label{eqnDCG}
 s_{\text{DCG}_{10\%}}(\brho) = \sum_{\ell=1}^L \frac{1}{\log_2(\ell + 1)}\IND{\rho^\ell \le 0.1 N}.
\end{equation}
The advantage of $p(\text{better than random})$ is that it can be averaged over seeds, folds, metrics, data sets, and models. Its disadvantage is that although it correlates with improvement over random search, it is not the same. It is possible that an engine consistently produces better sequences than random search, but the improvement is small.

\subsubsection{Score-based metrics}
\label{secScoreMetrics}

Score-based metrics answer the question: how much do we improve the score of random search by using a hyperopt engine? First, let us denote the test score of the best arm by $r^* = r^{\widehat{\rho}^1}$, where $\widehat{\rho}^1$ is the index of the arm with the best \emph{validation} score: $\widehat{\rho}^1 = \argmax_{\ell=1, \ldots, L} \widehat{r}^\ell$.\footnote{We assume here that the score is the higher, the better.} Note that, in general, $r^*$ is not the best test score $r^* \not= \max(r^1, \ldots, r^L)$ since $r^\ell \not= \widehat{r}^\ell$. The issue in using the numerical value of $r^*$ is that its scale depends on the model, the data set, and the metrics. To make this metrics easy to aggregate, we normalize it between $r^\text{rand}$, the expected best score of the random search with budget $L$, and $r^\text{grid} = r^{\widehat{\bi}^*}$, the test score of the best arm $\widehat{\bi}^* = \argmax_{\bi \in \cG} \widehat{r}^\bi$ in the full grid search, obtaining
\[
\widetilde{r}^* = 100 \frac{r^* - r^\text{rand}}{r^\text{grid} - r^\text{rand}}.
\]
$r^\text{rand}$ can be computed analytically by
\[
r^\text{rand} = \frac{\sum_{\ell=1}^{N} (1 - p)^{\ell - 1} r^{\widehat{\rho}_\ell}}{ \sum_{\ell=1}^{|\cT|} (1 - p)^{\ell - 1}},
\]
where $p = L / N$ is the probability of pulling a random arm, and $r^{\widehat{\rho}_\ell}$ is the test score of the $\ell$\/th rank statistics of the validation score table $\widehat{\cT}$.

Since we are interested in score improvement, when we aggregate $\widetilde{r}^*_i$ over a set of experiments $i$, we weight $\widetilde{r}^*_i$ by the maximum  possible improvement $\big(r^\text{grid}_i - r^\text{rand}_i\big)$, so, formally, the \emph{improvement degree} reported in Section~\ref{secResults} is given by
\begin{equation}\label{eqnImprovementDegree}
 \widetilde{r}^*_\text{mean} = 100 \frac{\sum_i \big(r^*_i - r^\text{rand}_i\big)}{\sum_i \big(r^\text{grid}_i - r^\text{rand}_i\big)}.
\end{equation}

\subsubsection{The overall score}

The baseline of $p(\text{better than random}) [\%]$ is 50, and the baseline of the improvement degree $\widetilde{r}^*$ is 0. The maximum of both scores is 100.\footnote{In fact the improvement degree can be larger than 100 in case grid search overfits.} We weight them equally, leading to
\begin{equation}\label{eqnOverall}
 \text{overall} = (p(\text{better than random}) [\%] - 50) + \widetilde{r}^* / 2.
\end{equation}

\section{Hyperopt engines}
\label{secEngines}

\href{https://docs.ray.io/en/latest/tune/index.html}{Ray Tune}~\citep{liaw2018} is a model selection and  hyperparameter optimization library that affords a unified interface to many popular hyperparameter engines. We used all engines ``out of the box'', according to the examples provided in the \href{https://docs.ray.io/en/latest/tune/examples/index.html#search-algorithm-examples}{documentation} (except for \href{https://sigopt.com}{\textsc{SigOpt}} -- behind paywall; and \href{https://github.com/dragonfly/dragonfly}{\textsc{Dragonfly}} -- cannot handle integer grid) to avoid ``overfitting'' our set of experiments. We also avoided consulting the authors to remain unbiased.  
\begin{enumerate}
    \item \href{https://ax.dev}{\textsc{\textcolor{ax}{AX}} \citep{bakshy2018}} is a domain-agnostic engine built and used by Meta for a wide variety of sequential optimization tasks (besides hyperopt, for A/B testing, infrastructure optimization, and hardware design). It links to  \textsc{BOTorch} \citep{balandat2020}, a Bayesian optimization library built on \textsc{GPyTorch} \citep{gardner2018}, a GP library built on \textsc{PyTorch} \citep{paszke2017}. It is one of the top three engines overall, and especially good in the low-number-of-trials regime, which may be due to the smart space-filling strategy that jump starts the optimization.

    \item \href{https://github.com/fmfn/BayesianOptimization}{\textsc{\textcolor{bayesopt}{BayesOpt}}} \citep{nogueira2014} is a standalone vanilla GP-based Bayesian optimization library. It performed very well on forest-type models, and very badly on SVM and neural nets.
    
    \item \href{https://github.com/automl/HpBandSter}{\textsc{\textcolor{bohb}{BOHB}}} \citep{falkner2018} combines Bayesian optimizaton and Hyperband \citep{li2018}, which is a bandit-based approach that speeds up random search using adaptive resource allocation and early stopping. In this study, \textsc{BOHB} did not beat random search, possibly due to the default settings, inadequate for our setup.
    
    \item \href{https://github.com/microsoft/FLAML/tree/main/flaml/tune#blendsearch-economical-hyperparameter-optimization-with-blended-search-strategy}{\textsc{\textcolor{cfo}{CFO}}} \citep{wu2021} is the first of two engines in Microsoft's \href{https://github.com/microsoft/FLAML}{FLAML} library. It is a local search method with randomized restarts. It is at a disadvantage in this study since its main forte is to manage trials with varying costs, whereas here we measure performance at a given number of trials.
    
    \item \href{https://github.com/microsoft/FLAML/tree/main/flaml/tune#blendsearch-economical-hyperparameter-optimization-with-blended-search-strategy}{\textsc{\textcolor{blend_search}{BlendSearch}}} \citep{wang2021} is the second of two engines in Microsoft's \href{https://github.com/microsoft/FLAML}{FLAML} library. It combines local search (CFO) with a global search. Its forte is still cost-sensitive optimization, and it uses no surrogate model, yet it is one of our top three engines overall. 
    
    \item \href{https://github.com/huawei-noah/HEBO/tree/master/HEBO}{\textsc{\textcolor{hebo}{HEBO}}} \citep{cowen2020} is Huawei's engine that won the \href{https://bbochallenge.com/}{2020 NeurIPS BBO Challenge}. It adds sophisticated processing steps to the classical BO framework, such as output warping, multi-objective acquisitions, non-stationary and heteroscedastic models, and input warping. It is one of the top three engines overall, and especially good with moderate and higher number of trials.
    
    \item \href{http://hyperopt.github.io/hyperopt/}{\textsc{\textcolor{hyperopt}{Hyperopt}}} \citep{bergstra2013} is one of the oldest hyperopt libraries, implementing \citet{bergstra2011}'s tree of Parzen estimators (TPE). It is tuned for high-dimensional computer vision tasks, so it seems to be disadvantaged in our low-dimensional low-number-of-trials regime.
    
    \item \href{https://github.com/facebookresearch/nevergrad}{\textsc{\textcolor{nevergrad}{Nevergrad}}} \citep{rapin2018} is a gradient-free optimization engine from Meta. As suggested by the Ray Tune documentation, we are using the 1+1 engine, which is perhaps not the best choice for our regime.
    
    \item \href{https://optuna.org/}{\textsc{\textcolor{optuna}{Optuna}}} \citep{akiba2019} is a new-generation hyperopt library providing features such as pruning and dynamically constructed search spaces. Here we use its basic Bayesian optimization core which barely beats random search.
    
    \item \href{https://scikit-optimize.github.io/stable/}{\textsc{\textcolor{skopt}{SkOpt}}} is an open-source community-developed Bayesian optimization library. It performs better than random search but overall does not reach the performance of the top engines. 
    
    \item \href{https://github.com/polixir/ZOOpt}{\textsc{\textcolor{zoopt}{ZOOpt}}} \citep{liu2018} is zeroth-order (derivative-free) optimization engine. It performs well on our ranking-based statistics but not on the score-based metrics. Its main issue seems to be that it uses only a small fraction of the trial budget: once it thinks it found the optimum, it stops exploring.
\end{enumerate}

\section{Results}
\label{secResults}

We tested all engines on five binary classification algorithms (Table~\ref{tabModels}) and five data sets (Table~\ref{tabData}). All data sets are downloaded from \href{https://www.openml.org/}{\textsc{OpenML}} \citep{vanschoren2013, feurer2019}. They were selected for their size, to be able to precisely estimate the expected score on the test set. Training and validation sizes are uniformly $4500$, and $500$, respectively. We found this size being the sweet spot between making the study meaningful and being able to run all the experiments\footnote{It took about two months to run all experiments on an 8CPU, 16GB RAM server, in semi-automatic mode (experiments need to be babysat due to the instabilities and minor bugs of the various libraries).}. In addition, the size of 5000 training points is also quite relevant to a lot of real-world applications, it is used in other horizontal studies on tabular data \citep{caruana2004}, and it is substantially larger than the data sets used in some other hyperopt studies \citep{cowen2020}. Experiments with 1000 training points (not reported) led to results that were not substantially different. Larger training and validation sets would mean smaller noise of the validation score, but it seems that the engines are not too sensitive to the noise level (relative standard errors are $\lessapprox 1\%$ on the validation score, $\lessapprox 0.1\%$ on the test score), so our results will likely hold with larger training samples.

\begin{table}[!ht]
  \caption{Data sets. All data sets are downloaded from \href{https://www.openml.org/}{\textsc{OpenML}} \citep{vanschoren2013, feurer2019}. Training and validation sizes are uniformly $4500$, and $500$, respectively. With the exception of \textsc{CoverType}, all data are binary classification; \textsc{CoverType} is multi-class, we use the two most populous classes here. $\text{AUC}^*$ is the best cross-validated test score obtained by full grid search over the five models (Table~\ref{tabModels}).}
  \label{tabData}
  \centering
  \begin{tabular}{llllll}
    \toprule
    Data set & $n_\text{trn}$ & $n_\text{val}$ & $n_\text{test}$ & \% majority & $\text{AUC}^*$\\
    \cmidrule(r){1-6}
    \textsc{BNGCreditG} & 4500 & 500 & 40222 & 75 & $0.9198$ \\
    \textsc{Adult} & 4500 & 500 & 995000 & 70 & $0.8397$ \\
    \textsc{CoverType} & 4500 & 500 & 490141 & 57 & $0.9103$ \\
    \textsc{Higgs} & 4500 & 500 & 93049 & 53 & $0.7735$ \\
    \textsc{Jannis} & 4500 & 500 & 62312 & 57 & $0.8419$ \\
    \bottomrule
  \end{tabular}
\end{table}

All data are binary classification with relatively balanced classes. We used the area under the ROC curve (AUC) as the target metric. Our experiments could and should be repeated on other tasks and metrics, but, similarly to the data size, unless the ``nature'' of the optimization problem (noise, smoothness) is radically different, our result should be generalizable. 

\begin{table}[!ht]
  \caption{Binary classification models. RF2: \href{https://scikit-learn.org/stable/modules/generated/sklearn.ensemble.RandomForestClassifier.html}{\textsc{ScikitLearn} \citep{pedregosa2011} random forests} with two hyperparameters; RF3: \href{https://scikit-learn.org/stable/modules/generated/sklearn.ensemble.RandomForestClassifier.html}{\textsc{ScikitLearn} \citep{pedregosa2011} random forests} with three hyperparameters; XGB: \href{https://xgboost.readthedocs.io/en/stable/}{\textsc{XGBoost}} \citep{chen2016}, SVM: \href{https://scikit-learn.org/stable/modules/generated/sklearn.svm.SVC.html}{\textsc{ScikitLearn} \citep{pedregosa2011} support vector machines}; PYTAB: \href{https://pytorch-tabular.readthedocs.io/en/latest/}{\textsc{PyTorch} Tabular} \citep{joseph2021}. $D$ is the number of hyperparameters, $N$ is the size of the search grid, $L$ is the minimum trial budget (we used $L, 2L, 3L$ budgets), and $\overline{\text{AUC}}$ is the mean of the best test scores over our experiments. More information and model-wise results are in Appendix~\ref{secAppendixResults}.}
  \label{tabModels}
  \centering
  \begin{tabular}{lllll}
    \toprule
    Model & $D$ & $N$ & $L = \sqrt{N}$ & $\overline{\text{AUC}}$
    \\
    \cmidrule(r){1-5}
    RF2 & 2 & $9 \times 7 = 63$ & 8 & $0.8496$ \\
    RF3 & 3 & $4 \times 9 \times 7 = 252$ & 16 & $0.8503$ \\
    XGB & 3 & $5 \times 9 \times 7 = 315$ & 18 & $0.8553$ \\
    SVM & 2 & $11 \times 10 = 110$ & 10 & $0.822$ \\
    PYTAB & 4 & $3 \times 3 \times 5 \times 3 = 135$ & 12 & $0.8248$ \\
\bottomrule
  \end{tabular}
\end{table}

We used five popular binary classification algorithms, designed specifically for small tabular data (Table~\ref{tabModels}). We evaluated the  rank-based metrics on folds independently, using ten seeds for each engine, ending up with a hundred runs for each (engine, model, data set) triple. We evaluated the score-based metrics on the cross-validated score (since this is what an experienced data scientist would do), using twenty-five seeds per triple. 

\begin{figure}[!ht]
  \centering
  \includegraphics[width=0.48\textwidth]{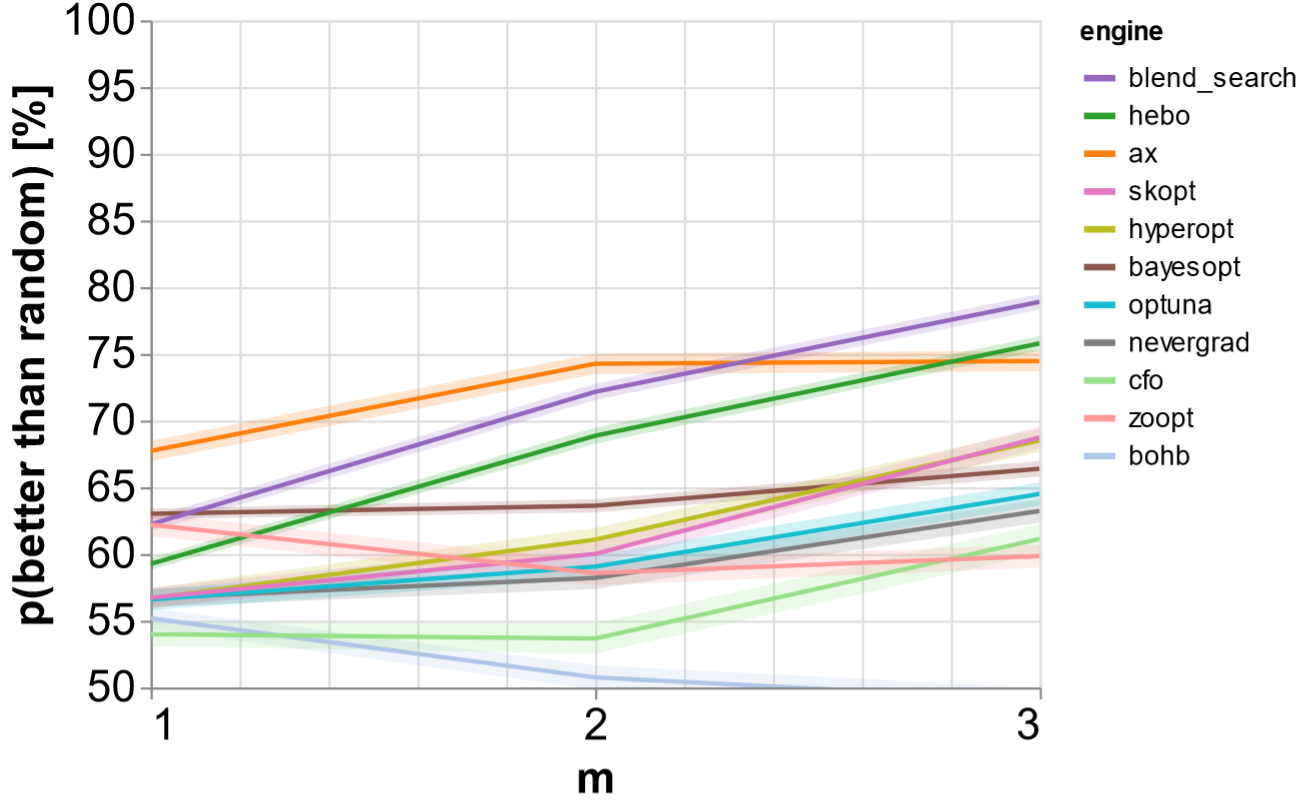}
  \includegraphics[width=0.48\textwidth]{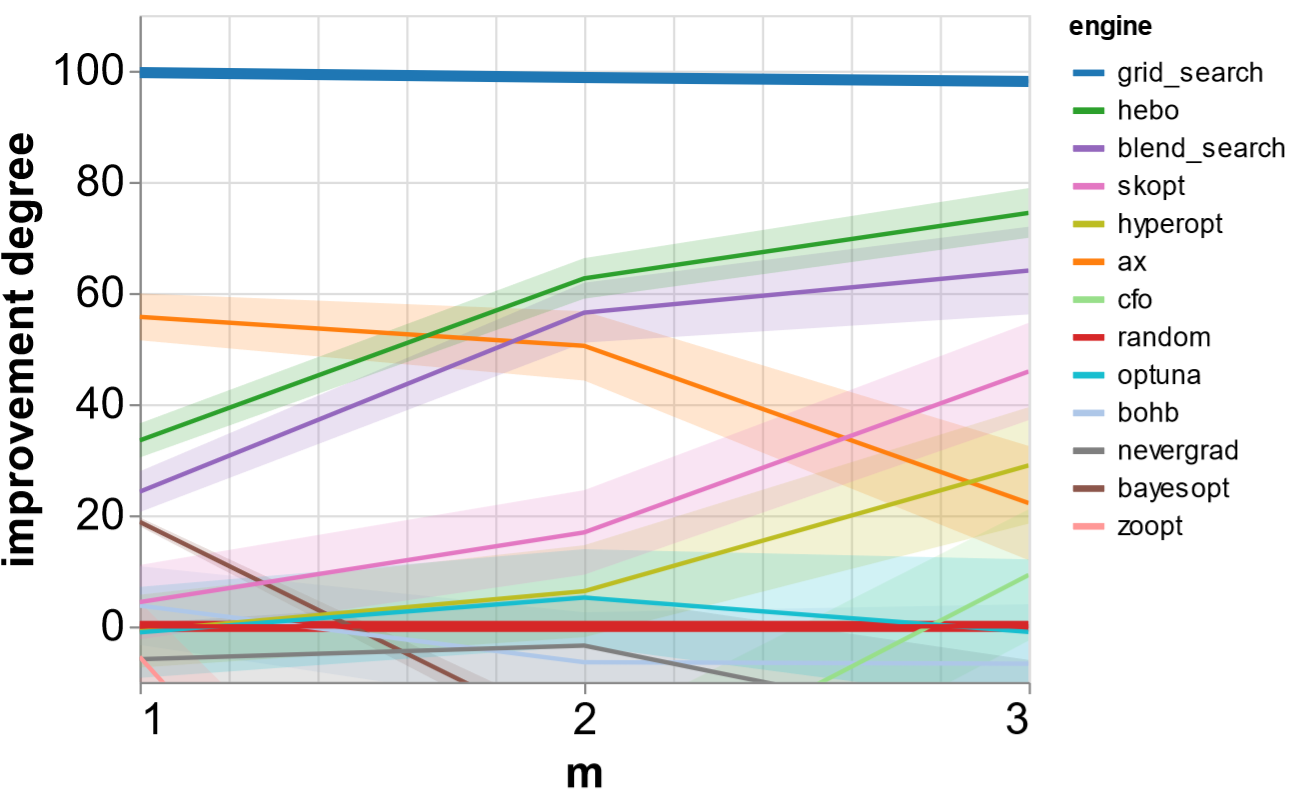}
  \caption{Summary of results. The probability that an engine is better than random (\ref{eqnBetterThanRandom}) is based on the DCG$_{10\%}$ statistics (\ref{eqnDCG}) computed on the arms pulled by the engine and $10^5$ draws of random search (Section~\ref{secRankMetrics}). Improvement degree (\ref{eqnImprovementDegree}) measures the improvement over random search, on a scale of 100, determined by the difference between the score of full grid search and random search (Section~\ref{secScoreMetrics}). Results are reported with three trial budgets $L = m  \times \sqrt{N}$, with $m=1, 2, 3$, where $N$ is the size of the full search grid.}
  \label{figResults}
\end{figure}

Table~\ref{tabResults} and Figure~\ref{figResults} show the rankings and the numerical results we obtained. All engines, except for \textsc{BOHB}, are significantly better than random search, although the gap varies between 55\% and 80\% probability (Table~\ref{tabResults}, columns 3-5). Three engines seem to perform significantly better than the rest: \textsc{HEBO}, \textsc{BlendSearch}, and \textsc{AX}. Using these engines, we can consistently accelerate random search by two to three times, or, from another angle, cut the difference between full grid search and random search with a given budget by about half (improvement degree = 50). \textsc{HEBO} is especially robust, managing to reach an improvement degree of 50 on all five models. The forte of \textsc{AX} is its performance on extra small budget ($m=1$). 

\begin{table}[!ht]
  \scriptsize
  \caption{Summary of results. The overall score is the sum of $(p(\text{better than random}) [\%] - 50)$ and  $0.5 \times \text{improvement degree}$. The probability that an engine is better than random (\ref{eqnBetterThanRandom}) is based on the DCG$_{10\%}$ statistics (\ref{eqnDCG}) computed on the arms pulled by the engine and $10^5$ draws of random search (Section~\ref{secRankMetrics}). Improvement degree (\ref{eqnImprovementDegree}) measures the improvement over random search, on a scale of 100, determined by the difference between the score of full grid search and random search (Section~\ref{secScoreMetrics}). Results are reported with three trial budgets $L = m  \times \sqrt{N}$, with $m=1, 2, 3$, where $N$ is the size of the full search grid. Forte is the set of models on which the engine has an improvement degree of 50 at any of the budgets.
  }
  \label{tabResults}
  \centering
  \begin{tabular}{lrr@{\;$\pm$\;}lr@{\;$\pm$\;}lr@{\;$\pm$\;}lr@{\;$\pm$\;}lr@{\;$\pm$\;}lr@{\;$\pm$\;}ll}
    \toprule
    Engine
    & overall
    & \multicolumn{6}{l}{$p(\text{better than random})$ $[\%]$}
    & \multicolumn{6}{l}{Improvement degree $[0, 100]$}
    & forte
    \\
    & score
    & \multicolumn{2}{c}{$m=1$}
    & \multicolumn{2}{c}{$m=2$}
    & \multicolumn{2}{c}{$m=3$}
    & \multicolumn{2}{c}{$m=1$}
    & \multicolumn{2}{c}{$m=2$}
    & \multicolumn{2}{c}{$m=3$}
    &
    \\
    \cmidrule(r){1-15}
    \color{hebo} \textsc{HEBO}
    & \color{hebo} 46
    &  59
    &  0
    &  69
    &  0
    &  76
    &  0
    &  33
    &  2
    &  \textbf{63}
    &  \textbf{2}
    &  \textbf{74}
    &  \textbf{3}
    &  RF2, RF3, SVM, XGB, PYTAB
    \\
    \color{blend_search} \textsc{BlendSearch}
    & \color{blend_search} 45
    &  62
    &  0
    &  72
    &  0
    &  \textbf{79}
    &  \textbf{0}
    &  24
    &  2
    &  56
    &  3
    &  64
    &  5
    &  RF2, RF3, SVM, XGB
    \\
    \color{ax} \textsc{AX}
    & \color{ax} 44
    &  \textbf{68}
    &  \textbf{0}
    &  \textbf{74}
    &  \textbf{0}
    &  74
    &  0
    &  \textbf{56}
    &  \textbf{3}
    &  50
    &  4
    &  22
    &  6
    &  RF2, RF3, XGB
    \\
    \color{skopt} \textsc{SkOpt}
    & \color{skopt} 23
    &  57
    &  0
    &  60
    &  1
    &  69
    &  0
    &  4
    &  4
    &  17
    &  5
    &  46
    &  5
    &  RF3, XGB
    \\
    \color{hyperopt} \textsc{Hyperopt}
    & \color{hyperopt} 18
    &  57
    &  0
    &  61
    &  1
    &  69
    &  1
    &  -1
    &  4
    &  6
    &  5
    &  29
    &  6
    &  XGB
    \\
    \color{optuna} \textsc{Optuna}
    & \color{optuna} 11
    &  57
    &  0
    &  59
    &  1
    &  65
    &  1
    &  -1
    &  5
    &  5
    &  5
    &  -1
    &  8
    &  
    \\
    \color{bayesopt} \textsc{BayesOpt}
    & \color{bayesopt} 6
    &  63
    &  0
    &  64
    &  0
    &  66
    &  0
    &  19
    &  0
    &  -21
    &  2
    &  -49
    &  6
    &  RF2, RF3, XGB
    \\
    \color{nevergrad} \textsc{Nevergrad}
    & \color{nevergrad} 5
    &  57
    &  0
    &  58
    &  1
    &  63
    &  1
    &  -6
    &  4
    &  -3
    &  6
    &  -20
    &  8
    &  SVM
    \\
    \color{bohb} \textsc{BOHB}
    & \color{bohb} 0
    &  55
    &  0
    &  51
    &  1
    &  49
    &  1
    &  4
    &  4
    &  -6
    &  5
    &  -7
    &  7
    &  
    \\
    \color{cfo} \textsc{CFO}
    & \color{cfo} -7
    &  54
    &  1
    &  54
    &  1
    &  61
    &  1
    &  -55
    &  9
    &  -32
    &  8
    &  9
    &  7
    &  
    \\
    \color{zoopt} \textsc{ZOOpt}
    & \color{zoopt} -43
    &  62
    &  1
    &  59
    &  1
    &  60
    &  1
    &  -5
    &  6
    &  -94
    &  12
    &  -217
    &  20
    &  
    \\
    \bottomrule
  \end{tabular}
\end{table}

The search grids we used are relatively coarse (Appendix~\ref{secAppendixResults}), and we found that this may disadvantage some of the engines, towards the bottom of the rankings. In preliminary experiments we found that some of these engines can pick up the difference if a finer grid is used. Nevertheless, the overall best result will not be better than when using a coarser grid, these engines improve only relatively to random search and to themselves with a coarser grid. Since the best engines in our rankings are more robust to grid resolution, even if one can afford a finer grid (see counterarguments in Section~\ref{secWhyGrid}), we suggest that our top three engines be used.

We found that some engines seem to specialize, for example \textsc{Nevergrad} is strong at optimizing SVM, whereas \textsc{SKOpt} is good at random forests and \textsc{XGBoost}. What this means is that in a comparison study of two algorithms on a data set, the winner may depend on which hyperopt engine is used. In fact, we found that out of the 8250 possible pairwise comparisons (pairs of models, pairs of engines, one of the data sets), about 4.3\% inverts the winner model. This may have quite serious consequences in systematic comparison studies, so in such studies we suggest that either full grid search or random search be used. This latter with add noise to the statistical tests but will not bias them.

\section{Conclusion and future works}

First, we are planning to repeat our methodology for other tasks (e.g., regression) and models. Second, some engines may have better settings for our coarse grid setup, so we are planning to design a protocol in which engine authors can give us a limited number of non-default settings to try. Third, we are planning to explore the effect of increasing the resolution of the grid, using our third protocol, to settle whether setting \emph{any} grid is solid advice or we should let all our search spaces as high-resolution as possible.

We paired our two metrics (rank-based and score-based) and two protocols (cross-validation and single fold) in two out of the four combinations. While most of the rankings match, there are curious differences which may be due to the metrics but also due to the noise level (which is about three times higher in the single validation case). We are planning to run a brief study to settle this question.

\bibliography{biblio}

\appendix

\section{Results for each model}
\label{secAppendixResults}

\newpage

\subsection{Random forests with two hyperparameters}

Hyperparameters and grid of values:
\begin{itemize}
    \item \texttt{max\_leaf\_nodes} $ = [2, 5, 10, 20, 50, 100, 200, 500, 1000]$
    \item \texttt{n\_estimators} $ = [10, 20, 50, 100, 200, 500, 1000]$
\end{itemize}

\begin{table}[!ht]
  \footnotesize
  \caption{Summary of results for random forests with two hyperparameters. The overall score is the sum of $(p(\text{better than random}) [\%] - 50)$ and  $0.5 \times \text{improvement degree}$. The probability that an engine is better than random (\ref{eqnBetterThanRandom}) is based on the DCG$_{10\%}$ statistics (\ref{eqnDCG}) computed on the arms pulled by the engine and $10^5$ draws of random search (Section~\ref{secRankMetrics}). Improvement degree (\ref{eqnImprovementDegree}) measures the improvement over random search, on a scale of 100, determined by the difference between the score of full grid search and random search (Section~\ref{secScoreMetrics}). Results are reported with three trial budgets $L = m  \times \sqrt{N}$, with $m=1, 2, 3$, where $N$ is the size of the full search grid.
  }
  \label{tabResultsRF2}
  \centering
  \begin{tabular}{lrr@{\;$\pm$\;}lr@{\;$\pm$\;}lr@{\;$\pm$\;}lr@{\;$\pm$\;}lr@{\;$\pm$\;}lr@{\;$\pm$\;}ll}
    \toprule
    Engine
    & overall
    & \multicolumn{6}{l}{$p(\text{better than random})$ $[\%]$}
    & \multicolumn{6}{l}{Improvement degree $[0, 100]$}
    \\
    & score
    & \multicolumn{2}{c}{$m=1$}
    & \multicolumn{2}{c}{$m=2$}
    & \multicolumn{2}{c}{$m=3$}
    & \multicolumn{2}{c}{$m=1$}
    & \multicolumn{2}{c}{$m=2$}
    & \multicolumn{2}{c}{$m=3$}
    \\
    \cmidrule(r){1-14}
    \color{ax} \textsc{AX}
    & \color{ax} 75
    &  66
    &  1
    &  \textbf{86}
    &  \textbf{1}
    &  \textbf{86}
    &  \textbf{1}
    &  \textbf{73}
    &  \textbf{4}
    &  \textbf{100}
    &  \textbf{0}
    &  \textbf{100}
    &  \textbf{0}
    \\
    \color{hebo} \textsc{HEBO}
    & \color{hebo} 49
    &  55
    &  0
    &  61
    &  1
    &  71
    &  1
    &  45
    &  0
    &  86
    &  4
    &  89
    &  2
    \\
    \color{blend_search} \textsc{BlendSearch}
    & \color{blend_search} 48
    &  61
    &  1
    &  68
    &  1
    &  75
    &  1
    &  8
    &  6
    &  81
    &  2
    &  92
    &  3
    \\
    \color{bayesopt} \textsc{BayesOpt}
    & \color{bayesopt} 34
    &  57
    &  0
    &  66
    &  0
    &  75
    &  1
    &  -37
    &  0
    &  68
    &  1
    &  76
    &  3
    \\
    \color{skopt} \textsc{SkOpt}
    & \color{skopt} 15
    &  60
    &  1
    &  50
    &  1
    &  63
    &  1
    &  -3
    &  9
    &  -23
    &  12
    &  70
    &  5
    \\
    \color{optuna} \textsc{Optuna}
    & \color{optuna} 11
    &  63
    &  1
    &  54
    &  1
    &  63
    &  1
    &  -3
    &  10
    &  -8
    &  14
    &  17
    &  17
    \\
    \color{bohb} \textsc{BOHB}
    & \color{bohb} 7
    &  63
    &  1
    &  53
    &  1
    &  51
    &  1
    &  15
    &  8
    &  -9
    &  13
    &  3
    &  14
    \\
    \color{hyperopt} \textsc{Hyperopt}
    & \color{hyperopt} -1
    &  62
    &  1
    &  49
    &  1
    &  58
    &  1
    &  -3
    &  9
    &  -60
    &  16
    &  21
    &  12
    \\
    \color{cfo} \textsc{CFO}
    & \color{cfo} -15
    &  62
    &  1
    &  52
    &  1
    &  54
    &  2
    &  -47
    &  16
    &  -76
    &  18
    &  -5
    &  20
    \\
    \color{zoopt} \textsc{ZOOpt}
    & \color{zoopt} -22
    &  \textbf{67}
    &  \textbf{1}
    &  65
    &  1
    &  65
    &  1
    &  12
    &  11
    &  -17
    &  23
    &  -221
    &  61
    \\
    \color{nevergrad} \textsc{Nevergrad}
    & \color{nevergrad} -24
    &  54
    &  0
    &  44
    &  1
    &  50
    &  1
    &  -36
    &  10
    &  -41
    &  13
    &  -65
    &  27
    \\
    \bottomrule
  \end{tabular}
\end{table}

\begin{figure}[!h]
  \centering
  \includegraphics[width=0.33\textwidth]{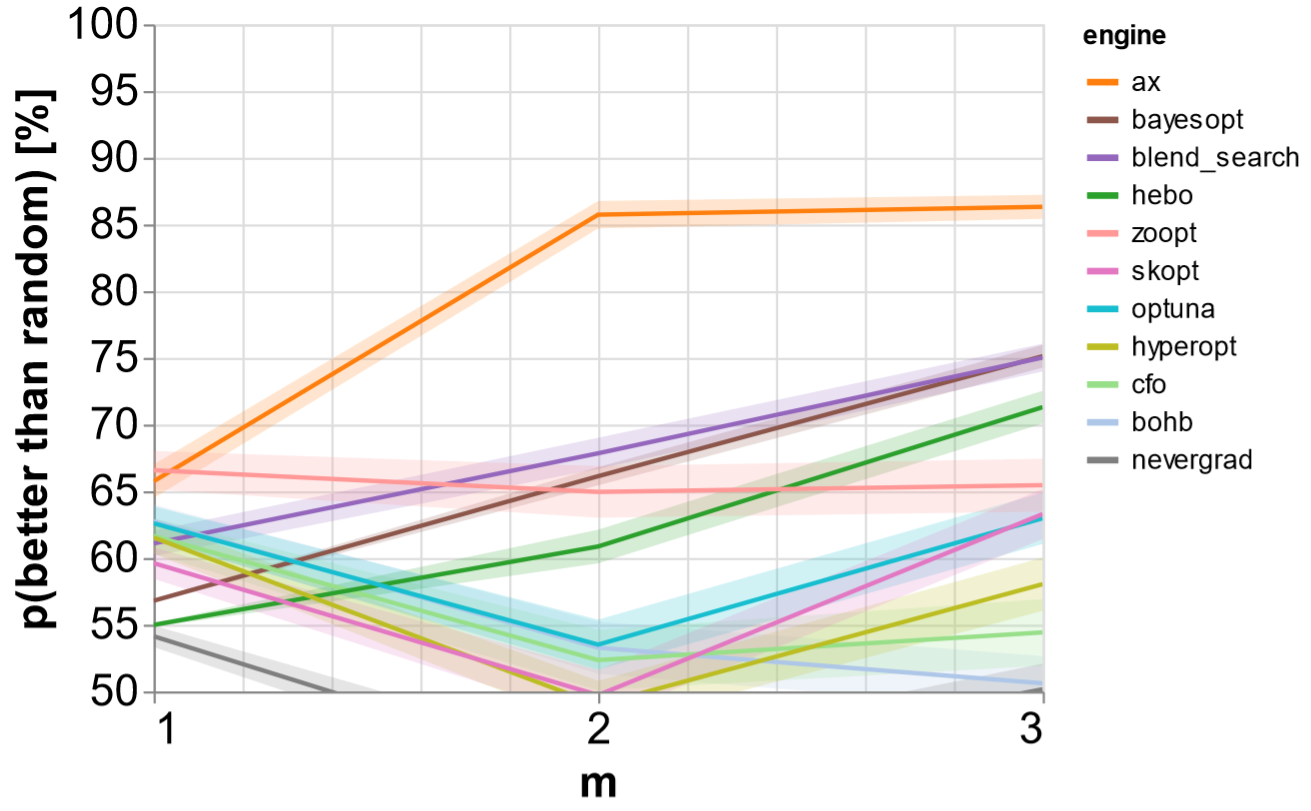}
  \includegraphics[width=0.33\textwidth]{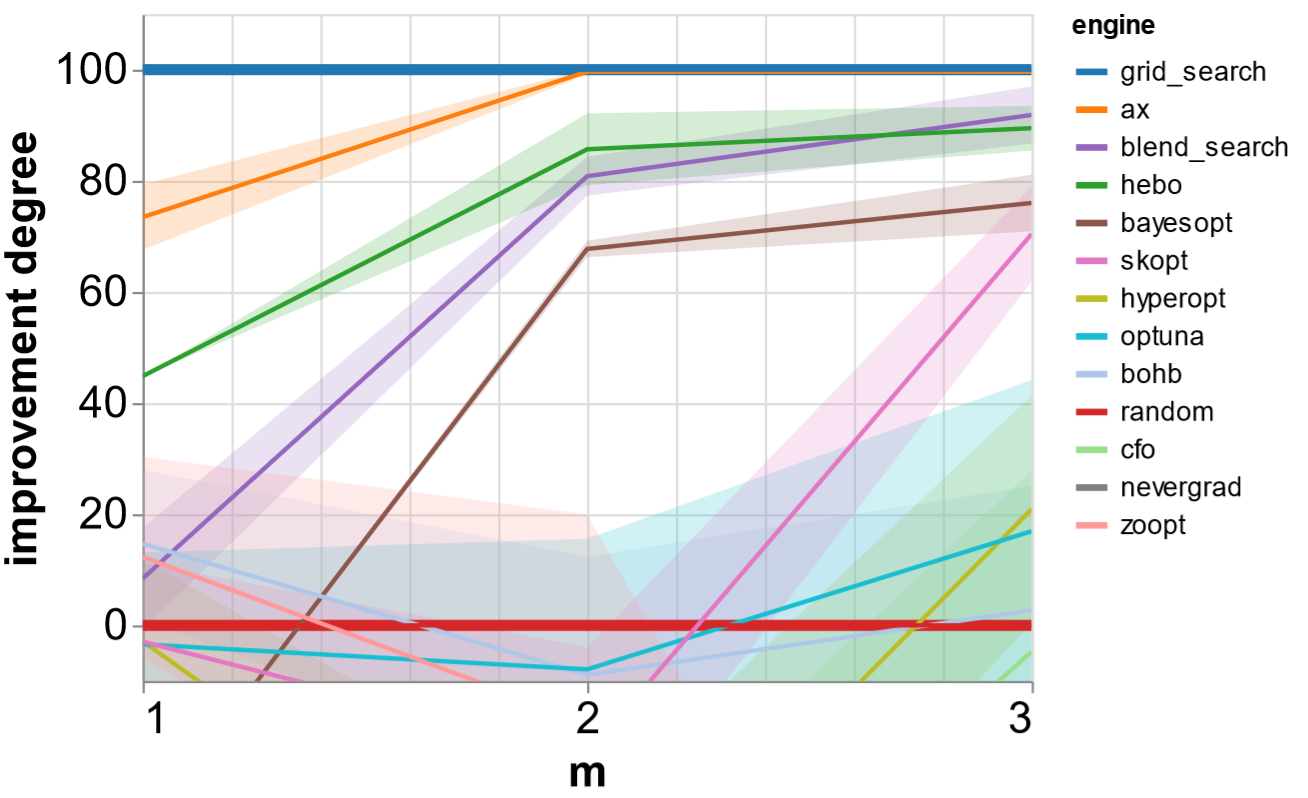}
  \caption{Summary of results for random forests with two hyperparameters. The probability that an engine is better than random (\ref{eqnBetterThanRandom}) is based on the DCG$_{10\%}$ statistics (\ref{eqnDCG}) computed on the arms pulled by the engine and $10^5$ draws of random search (Section~\ref{secRankMetrics}). Improvement degree (\ref{eqnImprovementDegree}) measures the improvement over random search, on a scale of 100, determined by the difference between the score of full grid search and random search (Section~\ref{secScoreMetrics}). Results are reported with three trial budgets $L = m  \times \sqrt{N}$, with $m=1, 2, 3$, where $N$ is the size of the full search grid.}
  \label{figResultsRF2}
\end{figure}

\newpage 

\subsection{Random forests with three hyperparameters}

Hyperparameters and grid of values:
\begin{itemize}
    \item \texttt{max\_features} $ = [0.1, 0.2, 0.5, 1.0]$
    \item \texttt{max\_leaf\_nodes} $ = [2, 5, 10, 20, 50, 100, 200, 500, 1000]$
    \item \texttt{n\_estimators} $ = [10, 20, 50, 100, 200, 500, 1000]$
\end{itemize}

\begin{table}[!ht]
  \footnotesize
  \caption{Summary of results for random forests with three hyperparameters. The overall score is the sum of $(p(\text{better than random}) [\%] - 50)$ and  $0.5 \times \text{improvement degree}$. The probability that an engine is better than random (\ref{eqnBetterThanRandom}) is based on the DCG$_{10\%}$ statistics (\ref{eqnDCG}) computed on the arms pulled by the engine and $10^5$ draws of random search (Section~\ref{secRankMetrics}). Improvement degree (\ref{eqnImprovementDegree}) measures the improvement over random search, on a scale of 100, determined by the difference between the score of full grid search and random search (Section~\ref{secScoreMetrics}). Results are reported with three trial budgets $L = m  \times \sqrt{N}$, with $m=1, 2, 3$, where $N$ is the size of the full search grid.
  }
  \label{tabResultsRF3}
  \centering
  \begin{tabular}{lrr@{\;$\pm$\;}lr@{\;$\pm$\;}lr@{\;$\pm$\;}lr@{\;$\pm$\;}lr@{\;$\pm$\;}lr@{\;$\pm$\;}ll}
    \toprule
    Engine
    & overall
    & \multicolumn{6}{l}{$p(\text{better than random})$ $[\%]$}
    & \multicolumn{6}{l}{Improvement degree $[0, 100]$}
    \\
    & score
    & \multicolumn{2}{c}{$m=1$}
    & \multicolumn{2}{c}{$m=2$}
    & \multicolumn{2}{c}{$m=3$}
    & \multicolumn{2}{c}{$m=1$}
    & \multicolumn{2}{c}{$m=2$}
    & \multicolumn{2}{c}{$m=3$}
    \\
    \cmidrule(r){1-14}
    \color{ax} \textsc{AX}
    & \color{ax} 81
    &  82
    &  1
    &  \textbf{95}
    &  \textbf{0}
    &  \textbf{93}
    &  \textbf{1}
    &  \textbf{81}
    &  \textbf{3}
    &  \textbf{88}
    &  \textbf{3}
    &  \textbf{77}
    &  \textbf{5}
    \\
    \color{bayesopt} \textsc{BayesOpt}
    & \color{bayesopt} 67
    &  \textbf{85}
    &  \textbf{0}
    &  93
    &  0
    &  90
    &  1
    &  74
    &  3
    &  56
    &  1
    &  38
    &  3
    \\
    \color{hebo} \textsc{HEBO}
    & \color{hebo} 60
    &  63
    &  1
    &  84
    &  1
    &  90
    &  0
    &  39
    &  5
    &  68
    &  3
    &  76
    &  5
    \\
    \color{skopt} \textsc{SkOpt}
    & \color{skopt} 45
    &  58
    &  1
    &  78
    &  1
    &  86
    &  1
    &  -9
    &  12
    &  65
    &  6
    &  71
    &  7
    \\
    \color{blend_search} \textsc{BlendSearch}
    & \color{blend_search} 37
    &  55
    &  1
    &  71
    &  1
    &  83
    &  1
    &  1
    &  8
    &  49
    &  6
    &  53
    &  6
    \\
    \color{hyperopt} \textsc{Hyperopt}
    & \color{hyperopt} 36
    &  55
    &  1
    &  74
    &  1
    &  86
    &  1
    &  6
    &  7
    &  30
    &  6
    &  49
    &  5
    \\
    \color{optuna} \textsc{Optuna}
    & \color{optuna} 31
    &  53
    &  1
    &  80
    &  1
    &  86
    &  1
    &  5
    &  9
    &  19
    &  9
    &  23
    &  8
    \\
    \color{nevergrad} \textsc{Nevergrad}
    & \color{nevergrad} 23
    &  63
    &  1
    &  77
    &  1
    &  88
    &  1
    &  -20
    &  11
    &  18
    &  7
    &  -12
    &  14
    \\
    \color{bohb} \textsc{BOHB}
    & \color{bohb} -7
    &  53
    &  1
    &  50
    &  1
    &  51
    &  1
    &  -8
    &  12
    &  -21
    &  15
    &  -24
    &  16
    \\
    \color{cfo} \textsc{CFO}
    & \color{cfo} -11
    &  50
    &  1
    &  60
    &  2
    &  71
    &  2
    &  -145
    &  27
    &  -18
    &  17
    &  36
    &  11
    \\
    \color{zoopt} \textsc{ZOOpt}
    & \color{zoopt} -26
    &  74
    &  1
    &  84
    &  1
    &  80
    &  1
    &  0
    &  14
    &  -106
    &  34
    &  -228
    &  54
    \\
    \bottomrule
  \end{tabular}
\end{table}

\begin{figure}[!h]
  \centering
  \includegraphics[width=0.33\textwidth]{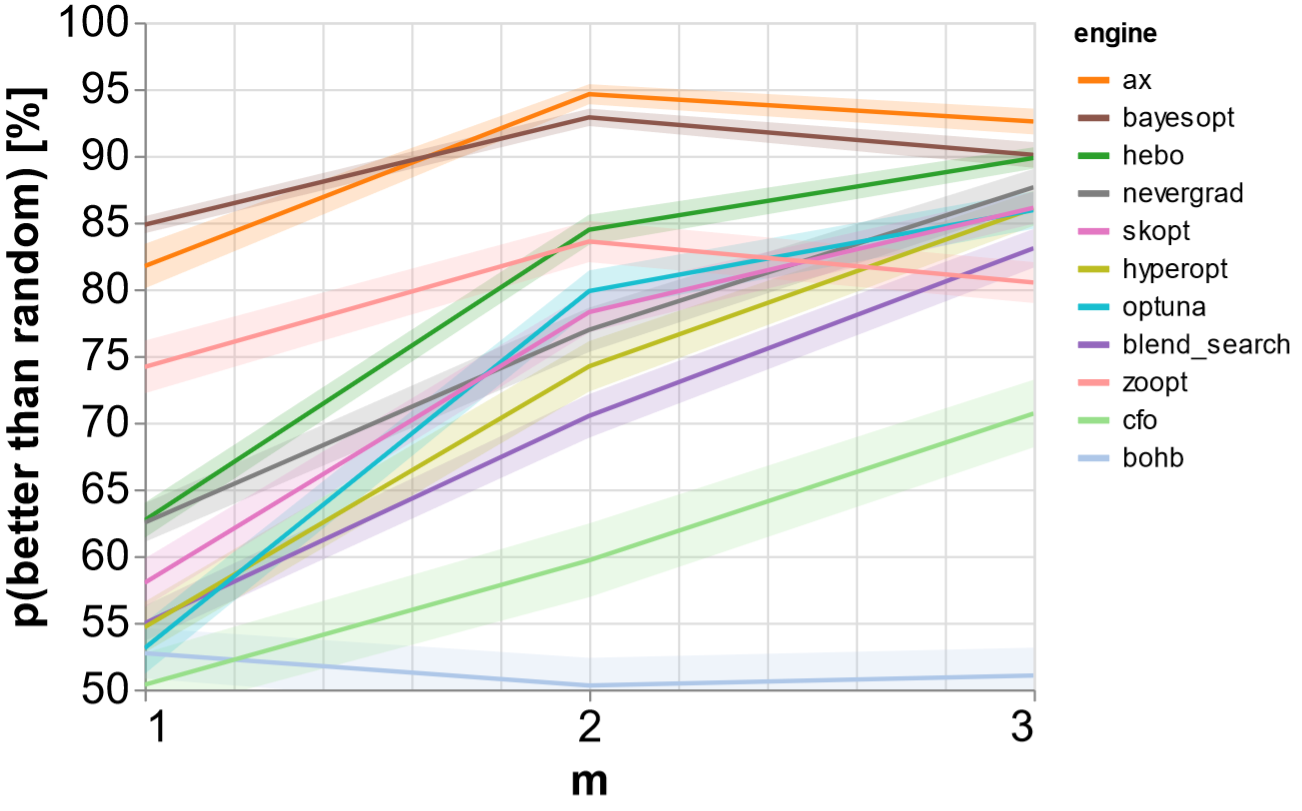}
  \includegraphics[width=0.33\textwidth]{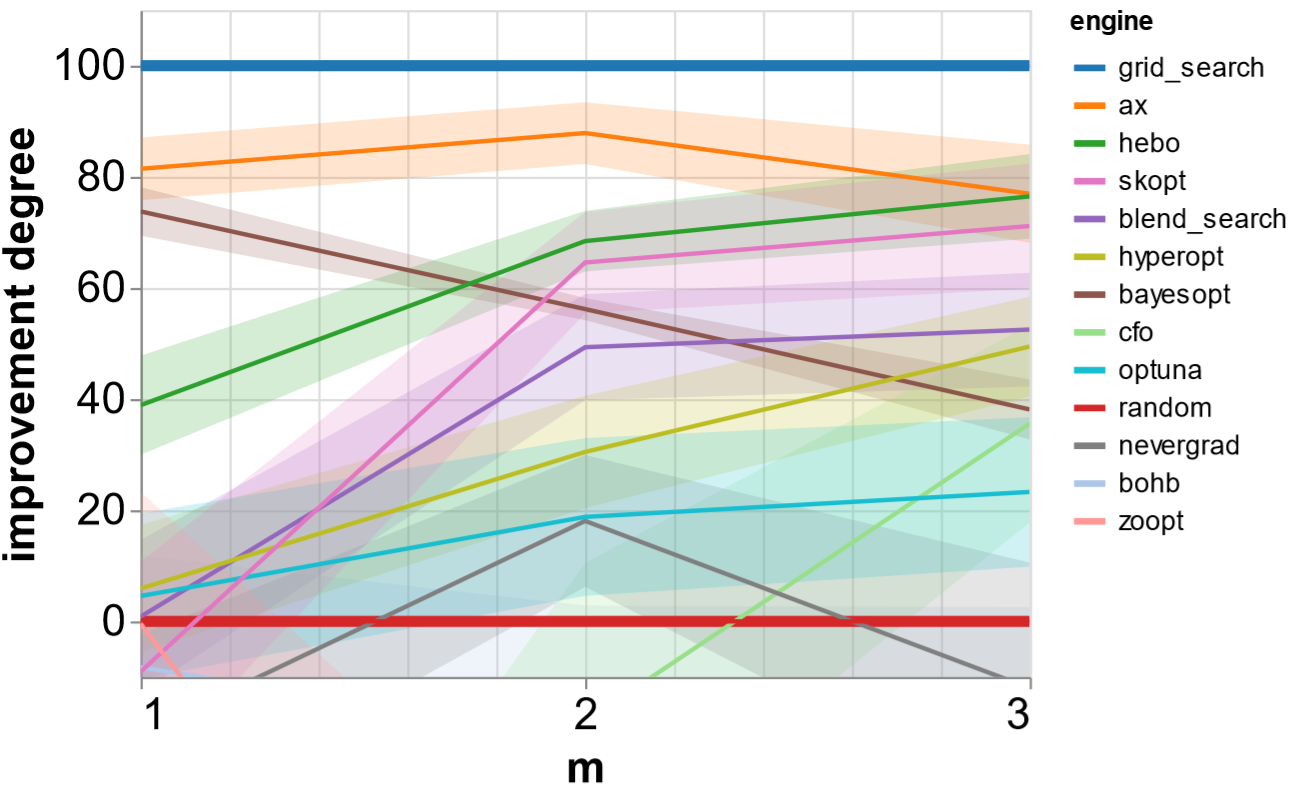}
  \caption{Summary of results for random forests with three hyperparameters. The probability that an engine is better than random (\ref{eqnBetterThanRandom}) is based on the DCG$_{10\%}$ statistics (\ref{eqnDCG}) computed on the arms pulled by the engine and $10^5$ draws of random search (Section~\ref{secRankMetrics}). Improvement degree (\ref{eqnImprovementDegree}) measures the improvement over random search, on a scale of 100, determined by the difference between the score of full grid search and random search (Section~\ref{secScoreMetrics}). Results are reported with three trial budgets $L = m  \times \sqrt{N}$, with $m=1, 2, 3$, where $N$ is the size of the full search grid.}
  \label{figResultsRF3}
\end{figure}

\newpage

\subsection{XGBoost with three hyperparameters}

Hyperparameters and grid of values:
\begin{itemize}
    \item \texttt{learning\_rate} $ = [0.1, 0.3, 0.5, 0.7, 1.0]$
    \item \texttt{max\_depth} $ = [2, 3, 4, 5, 7, 10, 20, 50]$
    \item \texttt{n\_estimators} $ = [10, 20, 50, 100, 200, 500, 1000]$
\end{itemize}

\begin{table}[!ht]
  \footnotesize
  \caption{Summary of results for XGBoost with three hyperparameters. The overall score is the sum of $(p(\text{better than random}) [\%] - 50)$ and  $0.5 \times \text{improvement degree}$. The probability that an engine is better than random (\ref{eqnBetterThanRandom}) is based on the DCG$_{10\%}$ statistics (\ref{eqnDCG}) computed on the arms pulled by the engine and $10^5$ draws of random search (Section~\ref{secRankMetrics}). Improvement degree (\ref{eqnImprovementDegree}) measures the improvement over random search, on a scale of 100, determined by the difference between the score of full grid search and random search (Section~\ref{secScoreMetrics}). Results are reported with three trial budgets $L = m  \times \sqrt{N}$, with $m=1, 2, 3$, where $N$ is the size of the full search grid.
  }
  \label{tabResultsXGB}
  \centering
  \begin{tabular}{lrr@{\;$\pm$\;}lr@{\;$\pm$\;}lr@{\;$\pm$\;}lr@{\;$\pm$\;}lr@{\;$\pm$\;}lr@{\;$\pm$\;}ll}
    \toprule
    Engine
    & overall
    & \multicolumn{6}{l}{$p(\text{better than random})$ $[\%]$}
    & \multicolumn{6}{l}{Improvement degree $[0, 100]$}
    \\
    & score
    & \multicolumn{2}{c}{$m=1$}
    & \multicolumn{2}{c}{$m=2$}
    & \multicolumn{2}{c}{$m=3$}
    & \multicolumn{2}{c}{$m=1$}
    & \multicolumn{2}{c}{$m=2$}
    & \multicolumn{2}{c}{$m=3$}
    \\
    \cmidrule(r){1-14}
    \color{bayesopt} \textsc{BayesOpt}
    & \color{bayesopt} 73
    &  \textbf{86}
    &  \textbf{0}
    &  \textbf{86}
    &  \textbf{1}
    &  80
    &  1
    &  \textbf{91}
    &  \textbf{1}
    &  79
    &  1
    &  66
    &  2
    \\
    \color{blend_search} \textsc{BlendSearch}
    & \color{blend_search} 71
    &  69
    &  1
    &  85
    &  1
    &  \textbf{90}
    &  \textbf{1}
    &  65
    &  2
    &  \textbf{85}
    &  \textbf{3}
    &  \textbf{88}
    &  \textbf{3}
    \\
    \color{hebo} \textsc{HEBO}
    & \color{hebo} 45
    &  52
    &  1
    &  76
    &  1
    &  \textbf{90}
    &  \textbf{1}
    &  0
    &  7
    &  54
    &  3
    &  80
    &  3
    \\
    \color{ax} \textsc{AX}
    & \color{ax} 39
    &  74
    &  1
    &  69
    &  1
    &  61
    &  2
    &  54
    &  6
    &  50
    &  7
    &  21
    &  13
    \\
    \color{hyperopt} \textsc{Hyperopt}
    & \color{hyperopt} 32
    &  50
    &  1
    &  70
    &  1
    &  74
    &  1
    &  4
    &  7
    &  51
    &  7
    &  50
    &  7
    \\
    \color{skopt} \textsc{SkOpt}
    & \color{skopt} 26
    &  51
    &  1
    &  62
    &  1
    &  70
    &  1
    &  8
    &  7
    &  34
    &  7
    &  45
    &  8
    \\
    \color{optuna} \textsc{Optuna}
    & \color{optuna} 14
    &  54
    &  1
    &  56
    &  1
    &  59
    &  1
    &  4
    &  7
    &  10
    &  10
    &  31
    &  11
    \\
    \color{nevergrad} \textsc{Nevergrad}
    & \color{nevergrad} 0
    &  46
    &  1
    &  52
    &  1
    &  51
    &  2
    &  -15
    &  9
    &  21
    &  10
    &  -2
    &  14
    \\
    \color{cfo} \textsc{CFO}
    & \color{cfo} -4
    &  47
    &  1
    &  51
    &  2
    &  61
    &  2
    &  -47
    &  13
    &  -27
    &  13
    &  31
    &  11
    \\
    \color{bohb} \textsc{BOHB}
    & \color{bohb} -5
    &  51
    &  1
    &  47
    &  1
    &  46
    &  1
    &  4
    &  9
    &  -8
    &  10
    &  -15
    &  12
    \\
    \color{zoopt} \textsc{ZOOpt}
    & \color{zoopt} -65
    &  52
    &  1
    &  37
    &  2
    &  30
    &  1
    &  7
    &  10
    &  -124
    &  21
    &  -214
    &  32
    \\
    \bottomrule
  \end{tabular}
\end{table}

\begin{figure}[!h]
  \centering
  \includegraphics[width=0.33\textwidth]{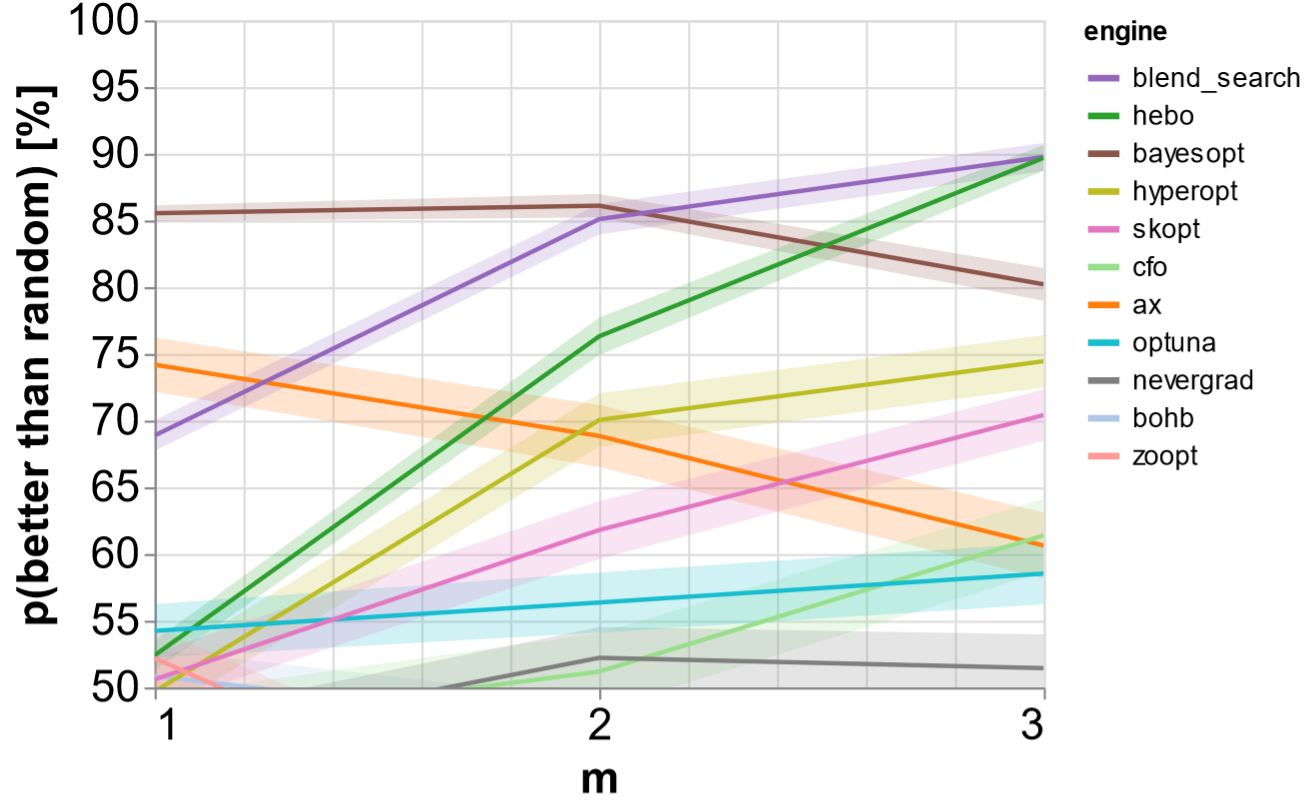}
  \includegraphics[width=0.33\textwidth]{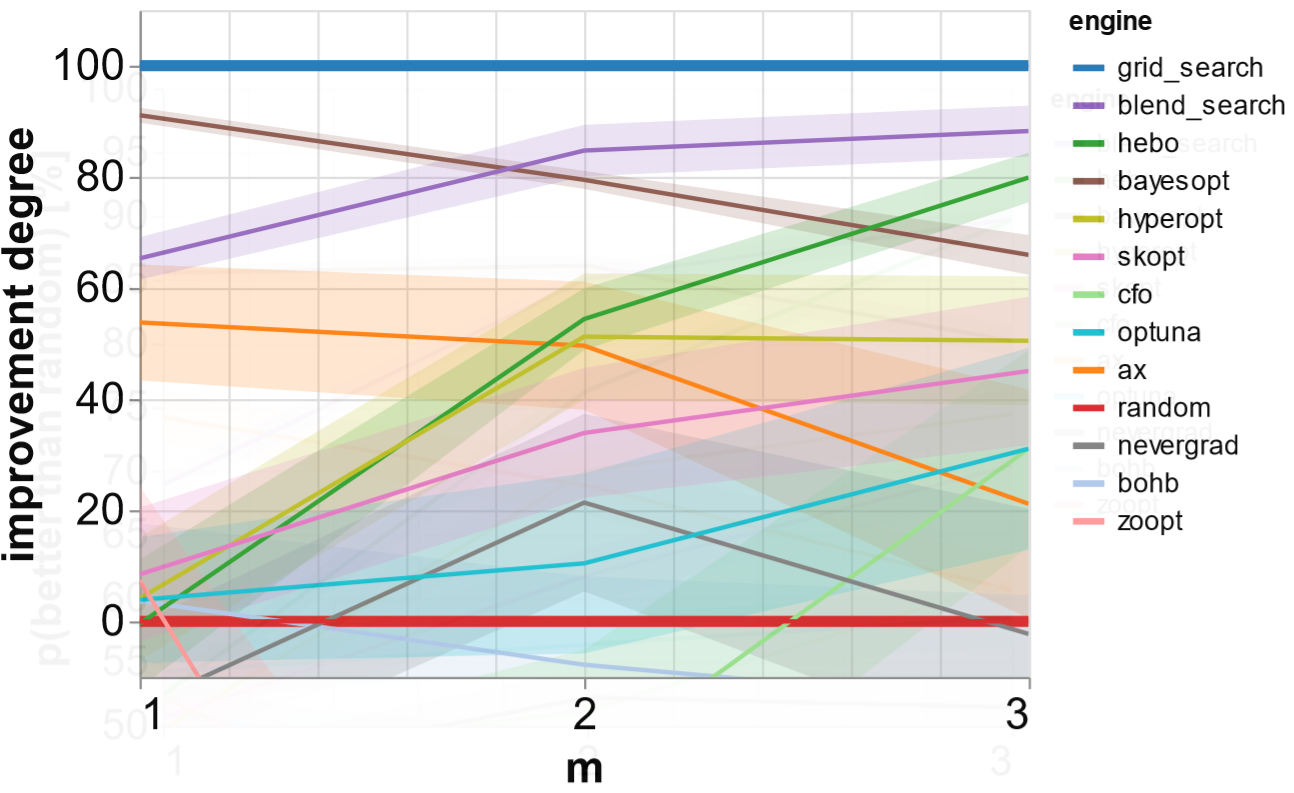}
  \caption{Summary of results for XGBoost with three hyperparameters. The probability that an engine is better than random (\ref{eqnBetterThanRandom}) is based on the DCG$_{10\%}$ statistics (\ref{eqnDCG}) computed on the arms pulled by the engine and $10^5$ draws of random search (Section~\ref{secRankMetrics}). Improvement degree (\ref{eqnImprovementDegree}) measures the improvement over random search, on a scale of 100, determined by the difference between the score of full grid search and random search (Section~\ref{secScoreMetrics}). Results are reported with three trial budgets $L = m  \times \sqrt{N}$, with $m=1, 2, 3$, where $N$ is the size of the full search grid.}
  \label{figResultsXGB}
\end{figure}

\newpage

\subsection{Support vector machines with two hyperparameters}

Hyperparameters and grid of values:
\begin{itemize}
    \item \texttt{C} $ = [0.03125, 0.125, 0.5, 2.0, 8.0, 32.0, 128.0, 512.0, 2.0480e+03, 8.1920e+03, 3.2768e+04]$
    \item \texttt{gamma} $ = [3.0518e-05, 0.0001221, 0.0004883, 0.001953, 0.007812, 0.03125, 0.125, 0.5, 2.0, 8.0]$
\end{itemize}

\begin{table}[!ht]
  \footnotesize
  \caption{Summary of results for support vector machines with two hyperparameters. The overall score is the sum of $(p(\text{better than random}) [\%] - 50)$ and  $0.5 \times \text{improvement degree}$. The probability that an engine is better than random (\ref{eqnBetterThanRandom}) is based on the DCG$_{10\%}$ statistics (\ref{eqnDCG}) computed on the arms pulled by the engine and $10^5$ draws of random search (Section~\ref{secRankMetrics}). Improvement degree (\ref{eqnImprovementDegree}) measures the improvement over random search, on a scale of 100, determined by the difference between the score of full grid search and random search (Section~\ref{secScoreMetrics}). Results are reported with three trial budgets $L = m  \times \sqrt{N}$, with $m=1, 2, 3$, where $N$ is the size of the full search grid.
  }
  \label{tabResultsSVM}
  \centering
  \begin{tabular}{lrr@{\;$\pm$\;}lr@{\;$\pm$\;}lr@{\;$\pm$\;}lr@{\;$\pm$\;}lr@{\;$\pm$\;}lr@{\;$\pm$\;}ll}
    \toprule
    Engine
    & overall
    & \multicolumn{6}{l}{$p(\text{better than random})$ $[\%]$}
    & \multicolumn{6}{l}{Improvement degree $[0, 100]$}
    \\
    & score
    & \multicolumn{2}{c}{$m=1$}
    & \multicolumn{2}{c}{$m=2$}
    & \multicolumn{2}{c}{$m=3$}
    & \multicolumn{2}{c}{$m=1$}
    & \multicolumn{2}{c}{$m=2$}
    & \multicolumn{2}{c}{$m=3$}
    \\
    \cmidrule(r){1-14}
    \color{hebo} \textsc{HEBO}
    & \color{hebo} 48
    &  66
    &  0
    &  65
    &  1
    &  68
    &  1
    &  \textbf{61}
    &  \textbf{2}
    &  59
    &  5
    &  \textbf{68}
    &  \textbf{10}
    \\
    \color{blend_search} \textsc{BlendSearch}
    & \color{blend_search} 46
    &  62
    &  1
    &  \textbf{70}
    &  \textbf{1}
    &  \textbf{78}
    &  \textbf{1}
    &  42
    &  4
    &  \textbf{62}
    &  \textbf{8}
    &  50
    &  15
    \\
    \color{nevergrad} \textsc{Nevergrad}
    & \color{nevergrad} 44
    &  \textbf{67}
    &  \textbf{1}
    &  67
    &  1
    &  73
    &  1
    &  49
    &  7
    &  41
    &  14
    &  55
    &  13
    \\
    \color{ax} \textsc{AX}
    & \color{ax} 31
    &  59
    &  1
    &  69
    &  1
    &  75
    &  1
    &  44
    &  6
    &  26
    &  9
    &  7
    &  15
    \\
    \color{hyperopt} \textsc{Hyperopt}
    & \color{hyperopt} 18
    &  60
    &  1
    &  55
    &  1
    &  65
    &  1
    &  -1
    &  11
    &  11
    &  13
    &  39
    &  17
    \\
    \color{skopt} \textsc{SkOpt}
    & \color{skopt} 13
    &  58
    &  1
    &  54
    &  1
    &  65
    &  1
    &  14
    &  9
    &  -3
    &  13
    &  11
    &  19
    \\
    \color{bohb} \textsc{BOHB}
    & \color{bohb} 1
    &  55
    &  1
    &  52
    &  1
    &  49
    &  1
    &  6
    &  11
    &  2
    &  12
    &  -16
    &  17
    \\
    \color{cfo} \textsc{CFO}
    & \color{cfo} -3
    &  58
    &  1
    &  56
    &  1
    &  59
    &  1
    &  -29
    &  23
    &  -21
    &  22
    &  -12
    &  19
    \\
    \color{optuna} \textsc{Optuna}
    & \color{optuna} -4
    &  58
    &  1
    &  52
    &  1
    &  61
    &  1
    &  -9
    &  16
    &  -7
    &  13
    &  -52
    &  25
    \\
    \color{zoopt} \textsc{ZOOpt}
    & \color{zoopt} -26
    &  60
    &  1
    &  60
    &  1
    &  65
    &  1
    &  -13
    &  16
    &  -50
    &  26
    &  -162
    &  36
    \\
    \color{bayesopt} \textsc{BayesOpt}
    & \color{bayesopt} -44
    &  36
    &  0
    &  27
    &  0
    &  38
    &  1
    &  31
    &  0
    &  -81
    &  8
    &  -118
    &  24
    \\
    \bottomrule
  \end{tabular}
\end{table}

\begin{figure}[!h]
  \centering
  \includegraphics[width=0.33\textwidth]{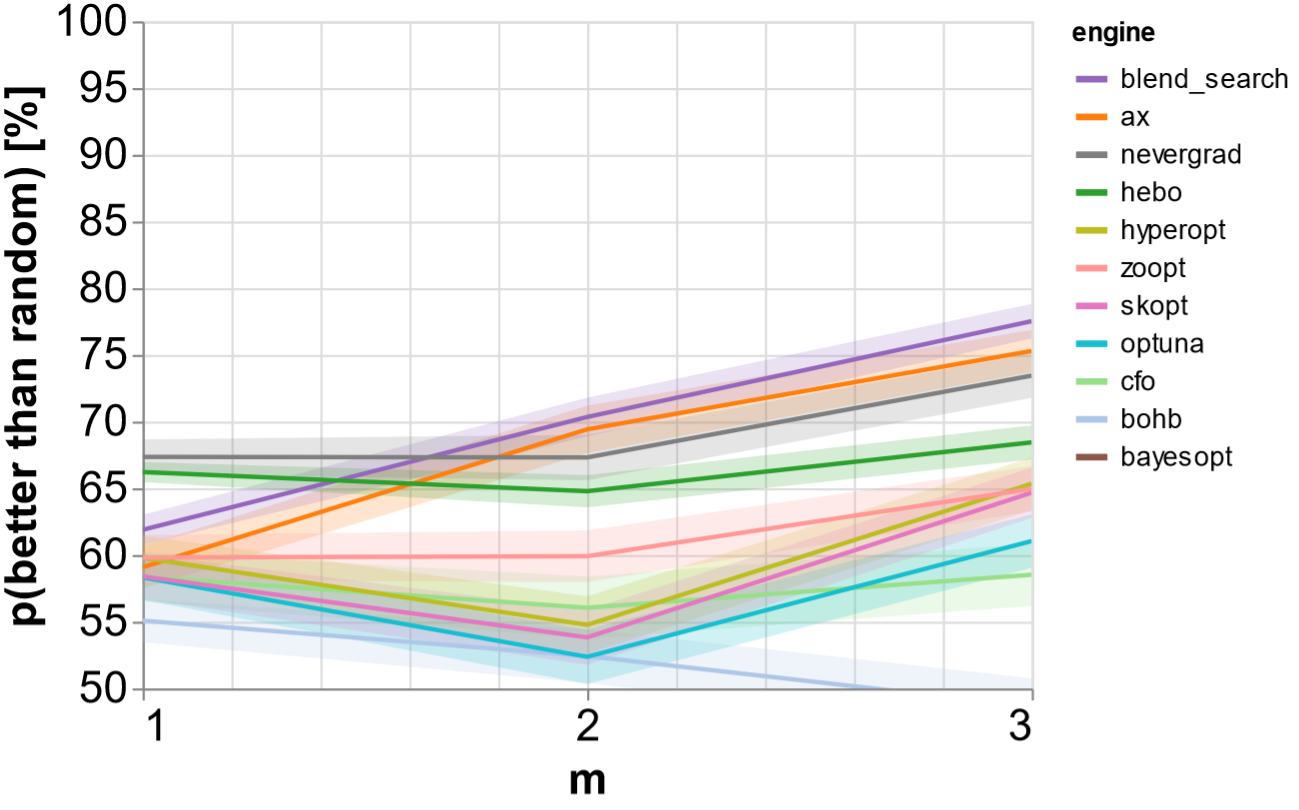}
  \includegraphics[width=0.33\textwidth]{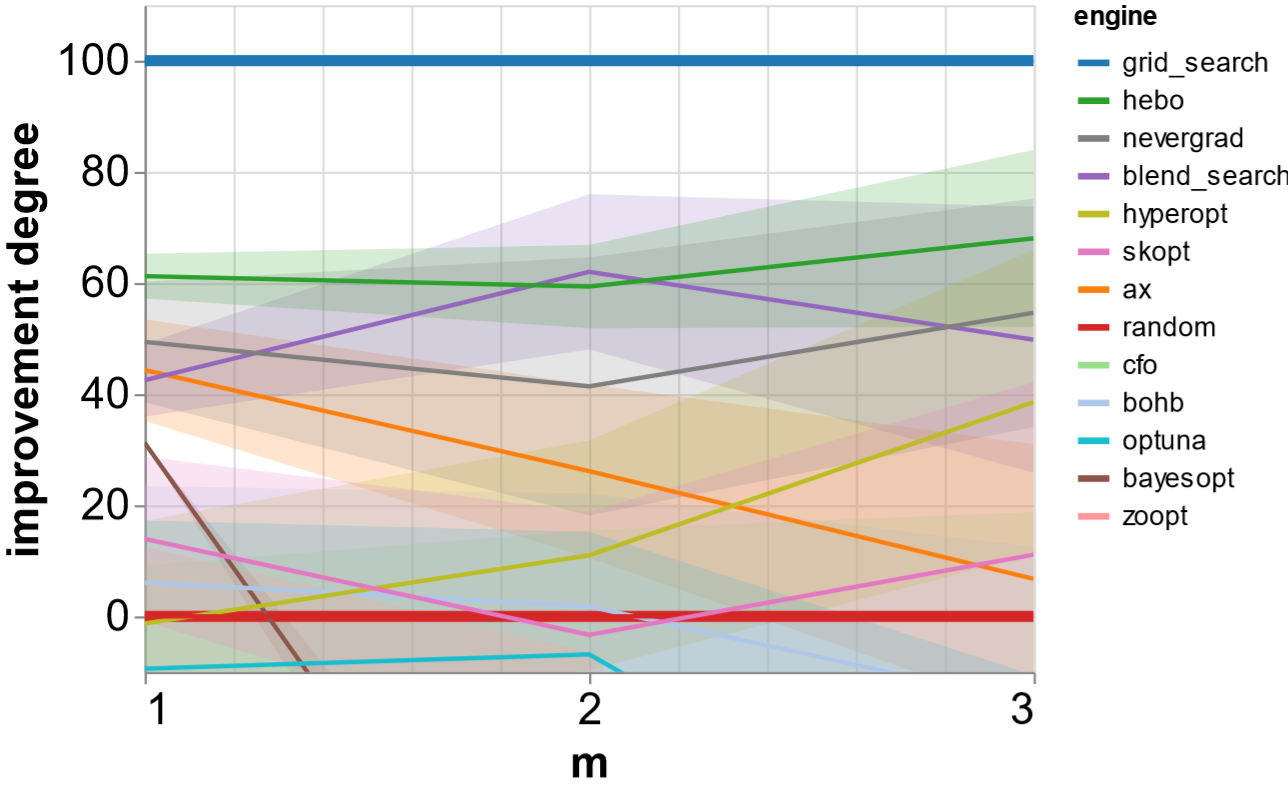}
  \caption{Summary of results for support vector machines with two hyperparameters. The probability that an engine is better than random (\ref{eqnBetterThanRandom}) is based on the DCG$_{10\%}$ statistics (\ref{eqnDCG}) computed on the arms pulled by the engine and $10^5$ draws of random search (Section~\ref{secRankMetrics}). Improvement degree (\ref{eqnImprovementDegree}) measures the improvement over random search, on a scale of 100, determined by the difference between the score of full grid search and random search (Section~\ref{secScoreMetrics}). Results are reported with three trial budgets $L = m  \times \sqrt{N}$, with $m=1, 2, 3$, where $N$ is the size of the full search grid.}
  \label{figResultsSVM}
\end{figure}

\newpage

\subsection{Pytab with four hyperparameters}

Hyperparameters and grid of values:
\vspace{-6pt}
\begin{itemize}
    \item \texttt{layers} $ = [512-256, 1024-512, 1024-512-512]$
    \item \texttt{learning\_rate} $ = [0.0001, 0.001, 0.01]$
    \item \texttt{n\_batches} $ = [1, 10, 20, 50, 100]$
    \item \texttt{n\_epochs} $ = [50, 100, 200]$
\end{itemize}
\vspace{-18pt}

\begin{table}[!ht]
  \footnotesize
  \caption{Summary of results for pytab with four hyperparameters. The overall score is the sum of $(p(\text{better than random}) [\%] - 50)$ and  $0.5 \times \text{improvement degree}$. The probability that an engine is better than random (\ref{eqnBetterThanRandom}) is based on the DCG$_{10\%}$ statistics (\ref{eqnDCG}) computed on the arms pulled by the engine and $10^5$ draws of random search (Section~\ref{secRankMetrics}). Improvement degree (\ref{eqnImprovementDegree}) measures the improvement over random search, on a scale of 100, determined by the difference between the score of full grid search and random search (Section~\ref{secScoreMetrics}). Results are reported with three trial budgets $L = m  \times \sqrt{N}$, with $m=1, 2, 3$, where $N$ is the size of the full search grid.
  }
  \label{tabResultsPYTAB}
  \centering
  \begin{tabular}{lrr@{\;$\pm$\;}lr@{\;$\pm$\;}lr@{\;$\pm$\;}lr@{\;$\pm$\;}lr@{\;$\pm$\;}lr@{\;$\pm$\;}ll}
    \toprule
    Engine
    & overall
    & \multicolumn{6}{l}{$p(\text{better than random})$ $[\%]$}
    & \multicolumn{6}{l}{Improvement degree $[0, 100]$}
    \\
    & score
    & \multicolumn{2}{c}{$m=1$}
    & \multicolumn{2}{c}{$m=2$}
    & \multicolumn{2}{c}{$m=3$}
    & \multicolumn{2}{c}{$m=1$}
    & \multicolumn{2}{c}{$m=2$}
    & \multicolumn{2}{c}{$m=3$}
    \\
    \cmidrule(r){1-14}
    \color{hebo} \textsc{HEBO}
    & \color{hebo} 30
    &  60
    &  1
    &  58
    &  1
    &  60
    &  1
    &  10
    &  3
    &  \textbf{52}
    &  \textbf{8}
    &  \textbf{62}
    &  \textbf{7}
    \\
    \color{blend_search} \textsc{BlendSearch}
    & \color{blend_search} 21
    &  \textbf{64}
    &  \textbf{1}
    &  \textbf{67}
    &  \textbf{1}
    &  \textbf{69}
    &  \textbf{1}
    &  -13
    &  4
    &  2
    &  12
    &  38
    &  17
    \\
    \color{skopt} \textsc{SkOpt}
    & \color{skopt} 19
    &  57
    &  1
    &  56
    &  1
    &  59
    &  1
    &  6
    &  9
    &  17
    &  11
    &  47
    &  13
    \\
    \color{bohb} \textsc{BOHB}
    & \color{bohb} 4
    &  55
    &  1
    &  51
    &  1
    &  49
    &  1
    &  -3
    &  9
    &  -1
    &  12
    &  22
    &  14
    \\
    \color{optuna} \textsc{Optuna}
    & \color{optuna} 4
    &  55
    &  1
    &  53
    &  1
    &  54
    &  1
    &  4
    &  8
    &  13
    &  12
    &  -19
    &  21
    \\
    \color{ax} \textsc{AX}
    & \color{ax} 3
    &  58
    &  1
    &  53
    &  1
    &  58
    &  1
    &  \textbf{30}
    &  \textbf{8}
    &  8
    &  14
    &  -59
    &  22
    \\
    \color{hyperopt} \textsc{Hyperopt}
    & \color{hyperopt} 0
    &  58
    &  1
    &  57
    &  1
    &  59
    &  1
    &  -10
    &  8
    &  -13
    &  12
    &  -21
    &  22
    \\
    \color{cfo} \textsc{CFO}
    & \color{cfo} -7
    &  53
    &  1
    &  49
    &  1
    &  61
    &  1
    &  -37
    &  13
    &  -24
    &  14
    &  -8
    &  19
    \\
    \color{nevergrad} \textsc{Nevergrad}
    & \color{nevergrad} -30
    &  54
    &  1
    &  51
    &  1
    &  53
    &  1
    &  -32
    &  8
    &  -67
    &  16
    &  -100
    &  25
    \\
    \color{zoopt} \textsc{ZOOpt}
    & \color{zoopt} -73
    &  58
    &  1
    &  47
    &  1
    &  59
    &  1
    &  -36
    &  15
    &  -164
    &  26
    &  -269
    &  46
    \\
    \color{bayesopt} \textsc{BayesOpt}
    & \color{bayesopt} -93
    &  52
    &  1
    &  46
    &  1
    &  48
    &  1
    &  -64
    &  0
    &  -206
    &  2
    &  -282
    &  18
    \\
    \bottomrule
  \end{tabular}
\end{table}

\begin{figure}[!h]
  \centering
  \includegraphics[width=0.33\textwidth]{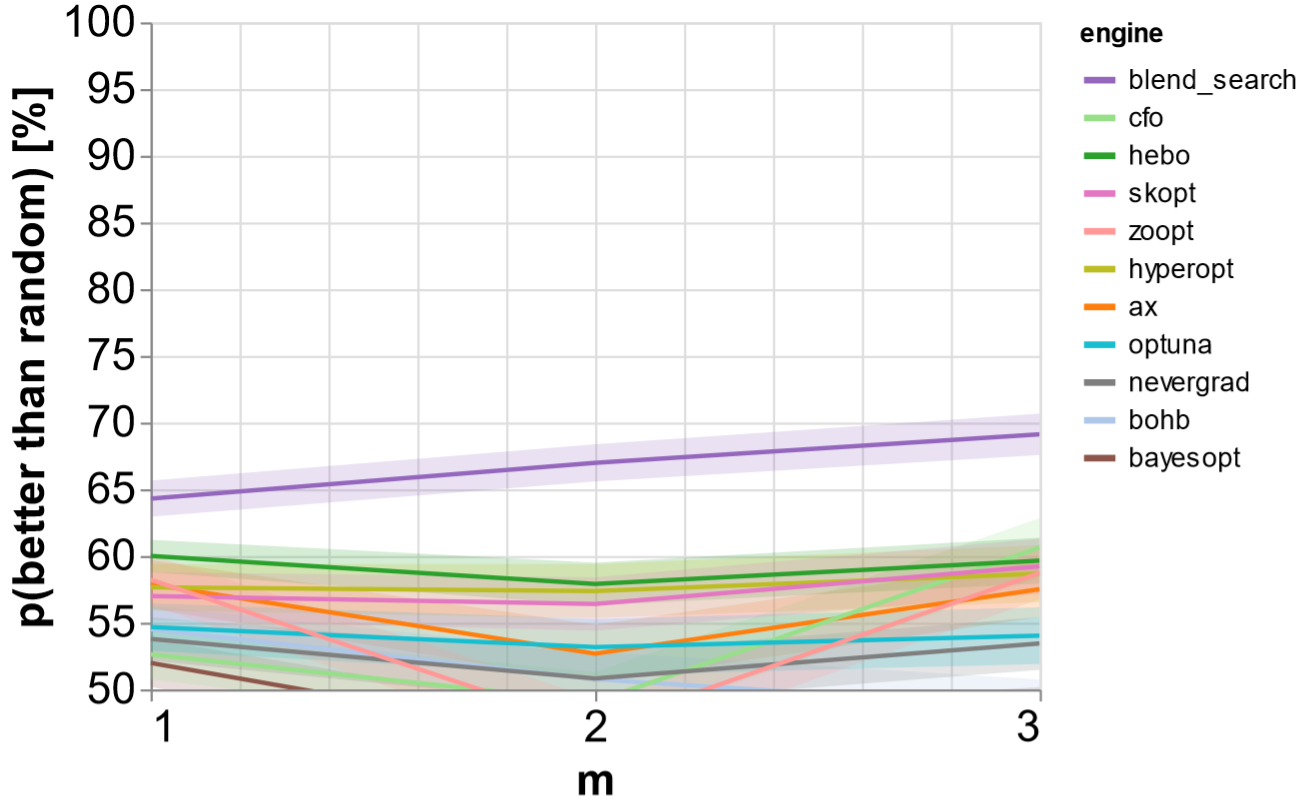}
  \includegraphics[width=0.33\textwidth]{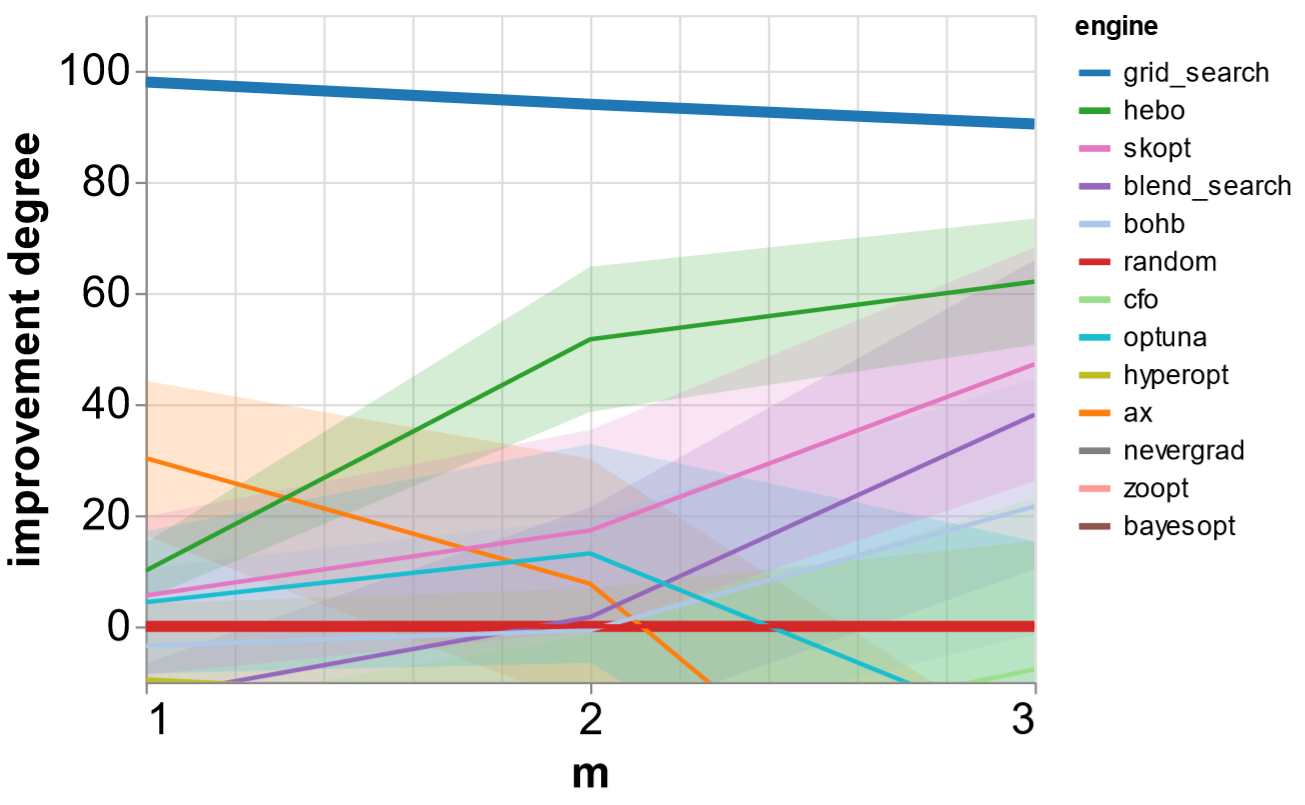}
  \caption{Summary of results for pytab with four hyperparameters. The probability that an engine is better than random (\ref{eqnBetterThanRandom}) is based on the DCG$_{10\%}$ statistics (\ref{eqnDCG}) computed on the arms pulled by the engine and $10^5$ draws of random search (Section~\ref{secRankMetrics}). Improvement degree (\ref{eqnImprovementDegree}) measures the improvement over random search, on a scale of 100, determined by the difference between the score of full grid search and random search (Section~\ref{secScoreMetrics}). Results are reported with three trial budgets $L = m  \times \sqrt{N}$, with $m=1, 2, 3$, where $N$ is the size of the full search grid.}
  \label{figResultsPYTAB}
\end{figure}

\end{document}